\documentclass[10pt,journal,compsoc]{IEEEtran}
\usepackage{amsmath}
\usepackage[mathscr]{euscript}
\usepackage{comment}
\usepackage{hyperref}
\hypersetup{
    colorlinks=true,
    linkcolor=blue,
    filecolor=blue,      
    urlcolor=blue,
    citecolor=blue,
}

\usepackage{ragged2e}

\usepackage{amsthm}
\newtheorem{definition}{Definition}

\def\,{\mskip\thinmuskip} \def\!{\mskip-\thinmuskip}

\ifCLASSOPTIONcompsoc
  \usepackage[nocompress]{cite}
\else
  \usepackage{cite}
\fi
\ifCLASSINFOpdf
\else
\fi

\usepackage{hyperref}

\usepackage{algorithm}
\usepackage[algo2e]{algorithm2e} 
\usepackage{algorithmicx}
\usepackage{textgreek}
\usepackage{textcomp}
\newcommand{\var}{\texttt}
\usepackage{xcolor}
\usepackage{graphicx}

\begin{document}

%
\title{A multi-schematic classifier-independent oversampling approach for imbalanced datasets}
%
%
%
%

\author{Saptarshi Bej, Kristian Schultz, Prashant Srivastava, Markus Wolfien,
        and Olaf Wolkenhauer
\IEEEcompsocitemizethanks{\IEEEcompsocthanksitem Saptarshi Bej, Kristian Schultz, Markus Wolfien, and Olaf Wolkenhauer are affiliated with the Faculty of Computer Science, University of Rostock, Germany 
\IEEEcompsocthanksitem  Olaf Wolkenhauer is affiliated by the Leibniz-Institute for Food Systems Biology, Technical University of Munich, Freising, 85354, Germany 
\IEEEcompsocthanksitem Prashant Srivastava is affiliated with Royal Holloway, University of London.

\IEEEcompsocthanksitem Olaf Wolkenhauer is the corresponding author of this paper \protect\\
Corresponding author address:\protect\\
  Department of Systems Biology \& Bioinformatics\protect\\
  University of Rostock, Universit\"atsplatz 1,\protect\\
  18051 Rostock, Germany\protect\\
  Tel.: +493814987570 \protect\\
  Fax: +493814987572\protect\\
  Orcid: 0000-0001-6105-2937\protect\\
  Email: olaf.wolkenhauer@uni-rostock.de

}
}

\IEEEtitleabstractindextext{%
\begin{abstract}
\justify
Labelled imbalanced data, used for classification problems, have an unequal distribution of samples over the classes. Traditional classification models, such as random forest, gradient boosting face a problem when dealing with imbalanced datasets. Over $85$ oversampling algorithms, mostly extensions of the SMOTE algorithm, have been built over the past two decades, to solve the problem of imbalanced datasets. 

However, it has been evident from previous studies that different oversampling algorithms have different degrees of efficiency with different classifiers. With numerous algorithms available, it is difficult to decide on an oversampling algorithm for a chosen classifier. Here we overcome this problem with a multi-schematic and classifier-independent oversampling approach, referred to as ProWRAS (Proximity Weighted Random Affine Shadowsampling). ProWRAS integrates the Localized Random Affine Shadowsampling (\href{https://link.springer.com/article/10.1007/s10994-020-05913-4}{LoRAS}) algorithm and the Proximity Weighted Synthetic oversampling (\href{https://link.springer.com/chapter/10.1007/978-3-642-37456-2_27}{ProWSyn}) algorithm.

By controlling the variance of the synthetic samples, as well as a proximity-weighted clustering system of the minority class data, the ProWRAS algorithm improves performance, compared to algorithms that generate synthetic samples through modelling high dimensional convex spaces of the minority class. ProWRAS is multi-schematic by employing four oversampling schemes, each of which has its unique way to model the variance of the generated data. The proximity weighted clustering approach of ProWRAS allows one to generate low variance synthetic samples only in borderline clusters to avoid overlap with the majority class. Most importantly, the performance of ProWRAS with proper choice of oversampling schemes, is independent of the classifier used.

We have benchmarked our newly developed ProWRAS algorithm against five sate-of-the-art oversampling models and four different classifiers on $20$ publicly available datasets. Our results show that ProWRAS outperforms other oversampling algorithms in a statistically significant way, in terms of both F1-score and $\kappa$-score. Moreover, we have introduced a novel measure for classifier independence $\mathscr{I}$-score, and showed quantitatively that ProWRAS performs better, independent of the classifier used. Thus, ProWRAS is highly effective for homogeneous tabular data where convex modelling of the data space can be done. In practice, ProWRAS customizes synthetic sample generation according to a classifier of choice and thereby reduce benchmarking efforts.

\end{abstract}

\begin{IEEEkeywords}
ProWRAS, LoRAS, SMOTE, Oversampling, Imbalanced datasets
\end{IEEEkeywords}}

\maketitle

\IEEEdisplaynontitleabstractindextext

%
\IEEEpeerreviewmaketitle

\section{Introduction}\label{sec:introduction}

Data originating from real-world problems are often imbalanced. Labelled imbalanced data, used for classification problems, have an unequal distribution of samples over the classes. The class(es) with a higher amount of samples is called majority class(es), and the class(es) with a smaller amount of samples is minority class(es).

Traditional Machine Learning based classification models, such as random forest, gradient boosting face certain difficulties, while dealing with such imbalanced datasets. In particular, due to the high number of majority samples encountered by the classifier, the learning gets biased towards the majority class. The minority class samples will have a higher chance of being mis-classified.

Over the years, several approaches, at data level (e.g., SMOTE, ADASYN) and algorithm level (e.g., cost-sensitive learning) have been developed to overcome the problems described above. Data level approaches involve pre-processing of data in a specific manner, and algorithm level approaches tend to modify various aspects of the classifiers in order to improve classification for imbalanced datasets. Oversampling approaches are data level approaches characterised by the generation of synthetic samples for the minority class in order to provide the classifiers with a balanced sample ratio. Finally, this leads to an improved learning experience for the minority class during the training process.

One of the most popular oversampling techniques that has been largely explored and researched upon is the Synthetic Minority Oversampling Technique (SMOTE)\cite{SMOTE}. SMOTE generates synthetic samples by generating random points along the line to join two close enough minority class samples, which can be interpreted as a convex combination of two close enough minority class samples.

However, the distribution of minority classes and latent noise in a data set is not taken into consideration by SMOTE \cite{MSMOTE}. Also, SMOTE over-generalises the minority class distribution, while generating synthetic samples, leading to classifiers biased towards minority class(es) \cite{Pruning, Blagus}.

To overcome such limitations, multiple extensions have been built as an improvement of SMOTE. These extensions implement a variety of approaches, such as the detection of borderline regions between classes and oversampling specifically from the borderline samples (Borderline-SMOTE and SVM SMOTE) and assigning sample weights to minority class samples to prioritise minority class samples to be used for synthetic sample generation (ADASYN and SMOTEBoost) \cite{Borderline-SMOTE, SVMSMOTE,ADASYN, SMOTEBoost, MWMOTE}. Other algorithms also detect clusters in the minority class to perform a prior learning of the minority class data distribution \cite{DBSMOTE, K-Means, MOT2LD, SOMO, CURESMOTE, LoRAS}. A recent study conducted an empirical comparison of $85$ such extensions or variants of the SMOTE algorithm proposed until $2018$ \cite{Comparison}. The comprehensive study benchmarked these algorithms on over a hundred imbalanced datasets using different classifiers, including Support Vector Machine (SVM), Decision Tree (DT), k-Nearest neighbours (kNN), and Multi Layered Perceptron (MLP), and investigated the best performing algorithms among the $85$ SMOTE extensions.

In 2020 Bej \textit{et al.}\ proposed the Localized Random Affine Shadowsampling (\href{https://link.springer.com/article/10.1007/s10994-020-05913-4}{LoRAS})) oversampling algorithm, which shows analytically that the local variance of the synthetic samples can be controlled by taking convex combinations of multiple shadowsamples (Gaussian noise added to minority class samples) from a minority class data neighbourhood, in contrast to taking convex combination of only two minority samples, as done by SMOTE and all of its prominent extensions \cite{LoRAS}. A benchmarking study on 14 publicly available datasets characterised by either of high-dimensionality, high-imbalance, and high-absolute imbalance, using three classifiers kNN, SVM, and Logistic Regression (LR), show that the approach of LoRAS improves the overall classification. However, from the empirical comparisons made by Kov\'acs, we notice that different oversampling approaches work well for different classifiers (See Table 3 in  \cite{Comparison}), although in Table 4, Kov\'acs, provided an aggregate ranking for the compared oversampling models \cite{Comparison}. According to this ranking, the two best algorithms overall are Polynom-fit SMOTE and ProWSyn.

Observing the dependence of oversampling algorithms on classifiers, we initially performed a pilot study, comparing the oversampling algorithms SMOTE, Polynom-fit SMOTE, ProWSyn, CURE-SMOTE, SOMO, and LoRAS using classifiers kNN, LR, Random Forest (RF), and Gradient Boosting (GB). The pilot study is a small scale study we perform before our main benchmarking experiment to verify our initial hypothesis about different oversampling models having varying efficiencies for different classifiers. We chose RF and GB as they are known to be powerful models using ensemble modelling approaches of Bagging and Boosting respectively \cite{GBRF}. Our study further confirmed that the performance of the oversampling algorithms are indeed classifier dependent. The question that motivated our research from this point was: 'Given that there are at least $85$ extensions of SMOTE for oversampling (and new ones being proposed every year), how does a user decide on which oversampling approach to use, considering that the performance of each of these oversampling approaches depends on the classifiers as well?'. Bej \textit{et al.} showed that the variance of the synthetic samples can affect the imbalanced data classification \cite{LoRAS}. Given a dataset and a classifier, it is difficult to decide the degree of variance of the synthetic samples that can lead to a better classification for the respective dataset.

As a solution, we developed a multi-schematic, classifier-independent oversampling approach, referred to as ProWRAS (Proximity Weighted Random Affine Shadowsampling). ProWRAS integrates the Localized Random Affine Shadowsampling (LoRAS) algorithm and the Proximity Weighted Synthetic oversampling (ProWSyn) algorithm. ProWRAS first creates partitions or clusters in the minority class data points, as per their proximity to the majority class. The clusters are then assigned normalised weights, such that clusters close to the majority class have higher weights. These weights decide the amount of synthetic samples to be generated from each cluster. To this point, the approach is similar to the ProWSyn algorithm \cite{ProWSyn}. 

The multi-schematic approach of ProWRAS conveniently generates synthetic samples with different degrees of variance using different oversampling schemes, providing us with a customized way to model the convex space, given a classifier and a dataset. ProWRAS thus overcomes the problem of laborious benchmarking studies to choose an appropriate oversampling algorithm from a pool of more than a hundred algorithms for a given dataset and a classifier of choice. Instead of benchmarking on tens of available algorithms, our results show that, a user can obtain good classifier performance by implementing only four oversampling schemes included in the ProWRAS approach. 

ProWRAS outperforms other oversampling algorithms in a statistically significant way, in terms of both F1-score and $\kappa$-score, independent of the classifier used. We benchmarked the ProWRAS algorithm on $20$ highly imbalanced, publicly available datasets. We employed four commonly used classifiers, GB, RF, kNN, and LR for this study.  Thus, ProWRAS is highly effective for homogeneous tabular data where convex modeling of the data space can be done. 


\begin{figure*}[!htbp]
\caption{Summarising oversampling schemes/strategies used by the investigated oversampling models and their respective influence on the classifier performance. For example, SMOTE generates samples with a ``High local variance'' scheme and works well for Gradient Boosting and Random Forest. Since the ProWRAS algorithm has access to all four oversampling schemes, its performance can be made independent of the chosen classifier.}
\label{PRoWRAS_motivation}
\centering
\includegraphics[scale=.99]{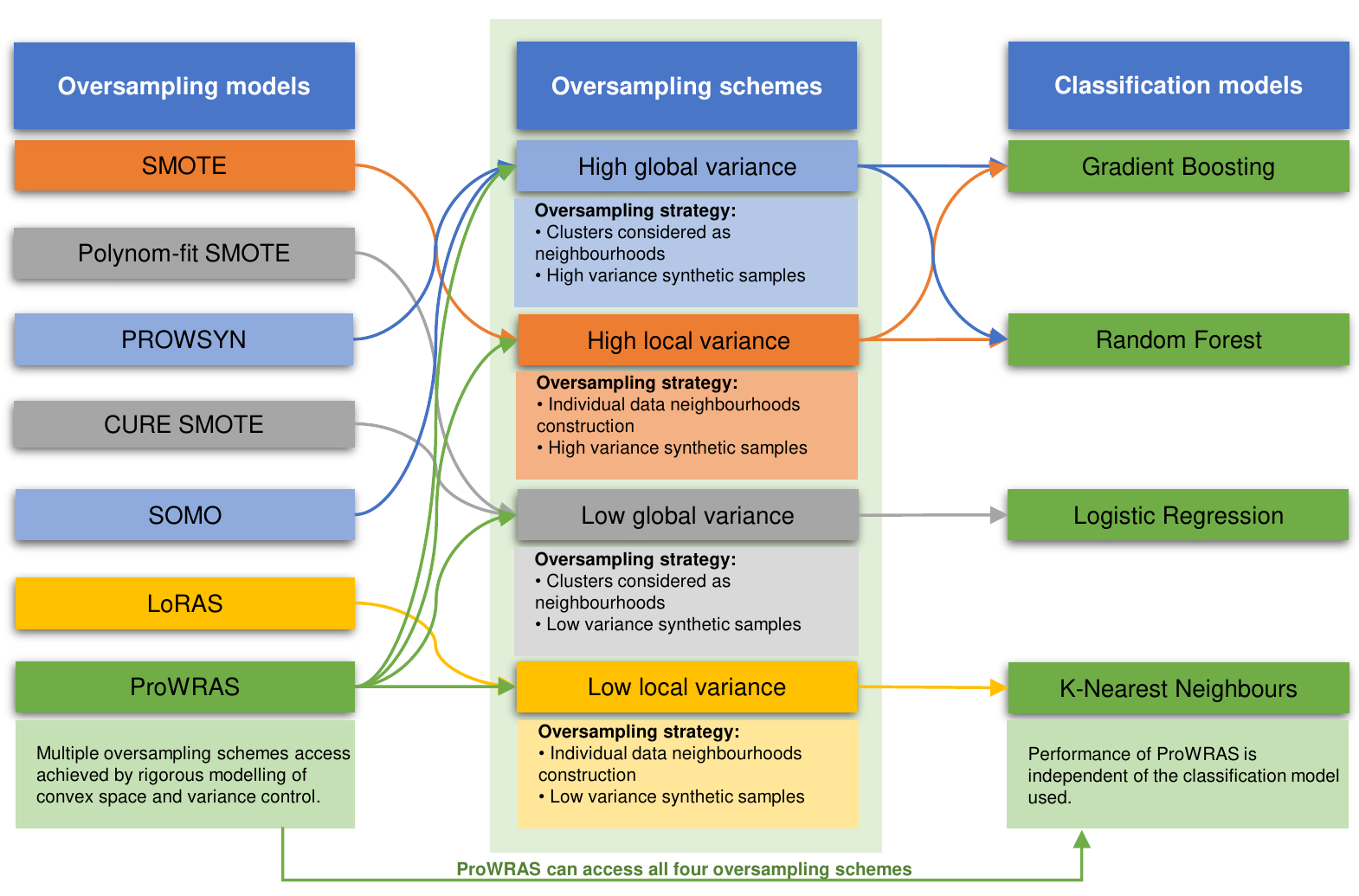}  
\end{figure*}


%
%
%
%

 



\section{Classifier Independent Oversampling}

\subsection{Quantification of classifier independence}

The basis of this work is the observation that the performance of existing variants of the well known SMOTE oversampling method for imbalanced classification problems are ``classifier dependent''. This is quite natural, as it is widely appreciated that, for machine learning, no single best method will exist with respect to all possible classification problems (the so-called no free-lunch theorem). Analogously, it is unlikely that there is a single best oversampling scheme, over all possible classifiers. However, this still poses problems. Since there are more than a hundred of such SMOTE variants and tens of ML based classifiers, given an imbalanced dataset, it is difficult to choose an appropriate oversampling algorithm from such a large pool of algorithms. We address this problem with the development of an oversampling approach, which offers good classification performance on an average, irrespective of the classifier used. In practice, this avoids laborious benchmarking experiments on numerous oversampling algorithms.\\

Some ambiguity may arise regarding the term `classifier independence 'since, an oversampling algorithm that always leads to worse than the others, could also be considered classifier independent, if its ranking is `stably' low. In order to establish some formalization about the term `classifier independence', we therefore propose a quantitative measure for classifier independence here. 

\begin{definition}
Given a set of oversampling algorithms $O$, a set of classifiers $C$ and a set of benchmarking datasets $D$, for a given oversampling algorithm $o \in O$, we define classifier independence of $o$ as,

\begin{equation}\label{CI}
\begin{split}
  \mathscr{I}(o)= \sqrt[\leftroot{1}\uproot{1}|C|]{\prod_{c\in C}\bigg(\frac{1}{|O|-1}\sum_{\substack{o'\in O\\o'\neq o}} \frac{\mathscr{F}(o',o)}{|D|} \bigg)}
\end{split}
\end{equation}
where, $\mathscr{F}(o',o)$, denotes the number of datasets for which the oversampling algorithm $o \in O$, performs equally or better than another oversampling algorithm $o' \in O$.  
\end{definition}

Note that, we choose a geometric mean over all classifiers in Equation \ref{CI}, to keep the measure strictly sensitive to classifier-specific performance of the oversampling algorithms, since the geometric mean is always less than the arithmetic mean. The value of $\mathscr{I}$, will always range between $0$ and $1$, making it a conveniently interpretable measure. $\mathscr{I}$ thus, measures not only how consistent the performance of an oversampling algorithm is over a set of classifiers, but also how well the oversampling model performs compared to other such models, overall.

\subsection{Algorithms used for benchmarking}\label{Algo}
As we have discussed in Section \ref{sec:introduction}, we benchmark the ProWRAS algorithm against SMOTE, Polynom-fit SMOTE, ProWSyn, CURE-SMOTE and LoRAS. We have used the SOMO algorithm in our pilot study but not in the main study because of its poor performance in terms of classifier-independene. In  Section \ref{Algo}, we first provide a brief description of the algorithms we used in our pilot study and benchmarking study, and then introduce the new ProWRAS algorithm. 

The \textbf{SMOTE} algorithm generates synthetic minority class samples by linear interpolation of the minority class. The synthetic sample generation approach is quite generic in its construct and focuses on the feature space of a given dataset. It has been described by \cite{SMOTE} using the following approach:

Let us assume that $x_1$ is an arbitrary minority class sample in an imbalanced dataset $C$. $C_\text{maj}$ and $C_ \text{min}$ are the majority and the minority class of $C$ respectively. Let us denote the set of $k$ the nearest minority class neighbours of $x_1$ by $N_k^{C_\text{min}}(x_1)$. We can also refer to $N_k^{C_\text{min}}(x_1)$ as the neighbourhood of the minority class data point $x_1$. Note that, for the sake of consistency, we will maintain these notations throughout. Let $x_2 \in N_k^{C_\text{min}}(x_1)$, $x_2 \neq x_1$ be a random minority class data point. A newly generated synthetic sample is described by:
\begin{equation}\label{SMOTE}
\begin{split}
   S= x_1+ u\cdot(x_2-x_1)
\end{split}
\end{equation}
where $0 < u < 1$. More samples can be generated from $N_k^{C_{min}}(x_1)$ simply by choosing additional random neighbours of $x_1$ within $N_k^{C_{min}}(x_1)$. Ultimately, this process can be repeated over all data points in the minority class to generate a population of synthetic samples from the minority class. Notably, Equation \ref{SMOTE}, is analogous to generating a convex combination of two minority class samples $x_1$ and $x_2$.
\begin{equation}\label{SMOTE_conv}
\begin{split}
   S= u\cdot x_1+ (1-u)\cdot x_2
\end{split}
\end{equation}

The \textbf{Polynom-fit SMOTE} (pf-SMOTE) algorithm was proposed in 2008 \cite{pfSMOTE}. This algorithm has different oversampling schemes based on underlying network topologies of the minority class. The pf-SMOTE algorithm proposes four different network topologies to generate synthetic samples from minority class, depending on the latent data distribution. These are: star topology, polynomial curve topology, bus topology, and mesh topology. The star topology generates synthetic samples along straight lines joining the mean of the minority class data points and each minority class data point, forming a star-like silhouette for the synthetic data. For polynomial curve topology, each feature is fit to a polynomial, the synthetic samples are generated feature-wise along the curve of these polynomials. For the bus topology, a path connecting one minority data to its nearest neighbour is formed using straight lines. Synthetic samples are sampled from this path. For the mesh topology, synthetic samples are generated along straight lines connecting each minority data point to the rest of the minority data points. The authors suggest that the star topology has proven to be the most effective \cite{pfSMOTE}. 

The \textbf{ProWSyn} algorithm partitions the minority class by their proximity to the majority class. The partitions are treated as clusters. The clusters are weighted as per their proximity to the majority class, such that clusters closer to the majority class have higher weights. The precise method for this is documented in Algorithm \ref{algoProWras:partitionInfo}. The weights decide how many samples are to be generated from each cluster. Synthetic samples are generated following the approach of SMOTE, that is, taking a convex combination of two arbitrary samples in a cluster. 

The CURE clustering algorithm is used by \textbf{CURE-SMOTE} to identify clusters among the minority class samples \cite{CURESMOTE}. The CURE algorithm uses a hierarchical clustering framework, which is also able to identify outliers in the minority class. The algorithm identifies centre points for each cluster after removing the noisy data points, in the process of clustering. It then generates synthetic data by applying SMOTE on each representative minority data point and the cluster centres \cite{CURESMOTE}.

The Self Organising Map Oversampling \textbf{SOMO} algorithm uses a Self Organising Map (SOM) \cite{SOM,SOMO}, a dimension reduction approach, to learn the latent distribution of the minority class data points. The imbalance ratio within each cluster is then calculated, followed by identification of clusters with an imbalance ratio of less than $1$. For each of these identified clusters a density factor quantifying the density of minority class data points in the respective cluster is calculated and then, every pair of topologically neighbouring clusters, another density factor, is calculated between the pairs. Both of the density factors are normalised to form a weight distribution. The weight distribution arising from the first distribution governs the amount of intra-cluster synthetic samples to be generated from each detected cluster, while the weight distribution arising from the second distribution governs the amount of inter-cluster synthetic samples to be generated between a pair of topologically close clusters \cite{SOMO}. The synthetic samples are generated using the SMOTE algorithm for both cases.
 
The \textbf{LoRAS} oversampling approach proposes to model the convex space more rigorously. Instead of generating synthetic samples by taking a convex combination of only two samples from a neighbourhood (as done by SMOTE and a majority of its extensions), LoRAS proposes to generate synthetic samples by taking convex combinations of multiple shadowsamples (Gaussian noise added to original minority class samples) in a minority class data neighbourhood\cite{LoRAS}.
Bej \textit{et al.}\ analytically calculates the variance of a LoRAS-generated synthetic sample (considered as a random variable) as:

\begin{equation}\label{LoRASvar}
\begin{split}
    \newcommand{\Var}{\mathrm{Var}}\Var(L_j) = \frac{2(\sigma'^2_j+\sigma_{Bj}^2)}{(|F|+1)} 
\end{split}
\end{equation}
where, $L_j$ is the $j$-th component of a LoRAS-generated sample $L$, $|F|$ is the number of shadowsamples considered for a convex combination to generate $L$, $\sigma'^2_j$ is the original variance of the minority class samples in a neighbourhood and $\sigma_{Bj}^2$ is the variance of the noise added to the original minority class samples to generate shadowsamples ($\sigma_{Bj}^2$ can be chosen to be arbitrarily small) \cite{LoRAS}. Moreover, the LoRAS algorithm uses manifold learning technique t-SNE to learn the minority class data neighbourhoods. 

The choice of algorithms is guided by previous benchmarking studies that identified the state-of-the-art. We include SMOTE because it is the pioneer of all algorithms. Polynom-fit SMOTE, ProWSyn were the top two oversampling algorithms by overall performance, from the detailed benchmarking study by Kov\'acs \cite{Comparison}. ProWSyn, CURE-SMOTE, and SOMO also use the idea of using clustering approaches on the minority class to learn the distribution of the minority class, a philosophy they share with ProWRAS algorithm. Finally, we chose LoRAS because it extended by the PrWRAS algorithm introduced in the next section. We further discuss the reasons for choosing these algorithms for benchmarking in Section \ref{Protocols for benchmarking}.

\subsection{ProWRAS algorithm}
\begin{algorithm*}[!htbp]
\small
    \SetArgSty{textnormal}
    \caption{ProWRAS oversampling algorithm (\href{https://github.com/Saptarshi-Bej/LoRAS-UMAP-project}{GitHub link})
    }
    \label{algo}
    \label{algoProWras}
    \SetKwInOut{In}{Inputs}
    \In{\newline
        \vspace{-1em}
        \begin{flushleft}
        \begin{tabular}{ c c }
            $\var{data}$ & Data points. \\ 
        \end{tabular}
        \end{flushleft}
    }
    \SetKwInOut{Parameter}{Parameters}
    \Parameter{
        \begin{flushleft}
        \vspace{-0.5em}
        \begin{tabular}{ l l l }
            $\var{max\_conv}$ 
            & $(> 0)$
            & Weight for number of generated samples per layer. 
            \\
            $\var{num\_samples\_to\_generate}$ 
            & $(> 0)$
            & Maximal count of generated samples in the output.
            \\
        \end{tabular}
        \end{flushleft}
    }

    
    \vspace{1em}

    \textbf{Function} \var{ProWRAS\_oversampling}(\var{data}) \Begin{
    $\var{clusters} \leftarrow{}$ \var{partition\_info(data)} \hspace*{.3cm} (See Algorithm \ref{algoProWras:partitionInfo}) 
    \vspace{3pt}

    $\var{weight\_max} \leftarrow{} max( \{ \var{weight} : (\var{cluster}, \var{weight}) \in \var{clusters} \} )$
    \vspace{3pt}

    Initialize $synth\_samples$ with an empty set.
    \vspace{5pt}

    \SetKwFor{For}{For}{do}{endfor}
    \SetKwIF{If}{ElseIf}{Else}{If}{then}{else If}{else}{endif}
    \For{$\var{(\var{cluster}, \var{weight}}) \in \var{clusters}$}{
        \vspace{5pt}
        $\var{num\_samples} \leftarrow{} \left\lceil \var{num\_samples\_to\_generate} \cdot \var{weight} \right\rceil$
        \vspace{5pt}

        $\var{num\_convcomb} \leftarrow{} \left\lceil{}\frac{\var{max\_conv} \cdot \var{weight}}{\var{weight\_max}}\right\rceil{}$
        \vspace{5pt}
    
        $\var{synth} \leftarrow{} 
        \var{generate\_points}(\var{cluster}, \var{num\_samples}, \var{num\_convcomb})$ \hspace*{.3cm} (See Algorithm \ref{algoProWras:generatePoints}) 
        \vspace{5pt}

        $\var{synth\_samples} \leftarrow{} \var{synth\_samples} \cup \var{synth}$
        \vspace{5pt}
    }

    Return resulting set of generated data points as $\var{synth\_samples}$.
    }
\end{algorithm*}

\begin{algorithm*}[!htbp]
\small
    \SetArgSty{textnormal}
        \caption{Proximity weighted minority class data partitioning}
    \label{algoProWras:partitionInfo}
    \SetKwInOut{In}{Inputs}
    \In{\newline
        \vspace{-1em}
        \begin{flushleft}
        \begin{tabular}{ c c }
            $\var{data}$ & Data points. \\ 
        \end{tabular}
        \end{flushleft}
    }
    \SetKwInOut{Parameter}{Parameters}
    \Parameter{
        \begin{flushleft}
    \begin{tabular}{ l l l }
        $\var{max\_levels}$ 
        & $(\geq 1)$ 
        & Maximal repeat of bordersearch. 
        \\
        $\var{n\_neighbours\_max}$ 
        & $(\geq 1)$ 
        & Number of neighbours considered for the majority class data points 
        \\
         
        & 
        & while constructing minority class partitions. 
        \\
        $\theta$ 
        & $(> 0)$ 
        & Scaling for weights. 
        \\
        $\var{num\_feats}$ 
        & $(= dim(x_1))$ 
        & Number of features. 
        \\
        \end{tabular}
        \end{flushleft}
    }

    
    \vspace{1em}

    \SetKwFor{For}{For}{do}{endfor}
    \SetKwIF{If}{ElseIf}{Else}{If}{then}{else If}{else}{endif}

    \textbf{Function} \var{partition\_info}(\var{data}) \Begin{

        $\var{X\_maj} \leftarrow{}$ Data points in $\var{data}$ with label for major class.

        $\var{X\_min} \leftarrow{}$ Data points in $X$ with label for minor class.
        \vspace{3pt}
        
        $L=\var{max\_levels}$
        
        Initialize $\var{clusters}$ as empty set.
        \vspace{3pt}

        \For{$i = 1 , 2, \ldots, L-1$}{
            \If{$|\var{X\_min}| = 0$}{break}
            
            $\var{weight} = \exp(- \theta \cdot (i-1))$
            \vspace{3pt}

            $k \leftarrow{} \min(|\var{X\_min}|, \var{n\_neighbours\_max]})$

            $\var{cluster} \leftarrow{}$ All neighbours in $k$-neighbourhoods from \var{X\_maj} in $\var{X\_min}$
            \vspace{3pt}

            $\var{clusters} = \var{clusters} \cup \{ (\var{cluster}, \var{weight}) \}$
            \vspace{3pt}

            $\var{X\_min} \leftarrow{} \var{X\_min} \setminus \var{cluster}$
        }
        \vspace{3pt}

        \If{$|\var{X\_min}| > 0$}{
            $\var{weight} = \exp(- \theta \cdot (L - 1))$

            $\var{clusters} = \var{clusters} \cup \{ (\var{X\_min}, \var{weight}) \}$
        }
        \vspace{3pt}
        $\var{weight\_sum} \leftarrow{} sum( \{ \var{weight} : (\var{cluster}, \var{weight}) \in \var{clusters} \} )$

        \vspace{3pt}
        $\var{clusters} \leftarrow{} \left\{ \left(\var{cluster}, \frac{\var{weight}}{\var{weight\_sum}}\right) : (\var{cluster}, \var{weight}) \in \var{clusters} \right\}$

        \vspace{3pt}
        Returns pairs of clusters and normalized weights as $\var{clusters}$.
        
        }
\end{algorithm*}

\begin{algorithm*}[!htbp]
\small
    \SetArgSty{textnormal}
        \caption{Cluster-wise oversampling schemes}
    \label{algoProWras:generatePoints}
    \SetKwInOut{In}{Inputs}
    \In{\newline
        \vspace{-1em}
        \begin{flushleft}
        \begin{tabular}{ l l }
            $\var{cluster}$ 
            & Data points. 
            \\ 
            $\var{num\_samples}$ 
            & Number of generated shadowsamples per parent data point. 
            \\ 
            $\var{num\_convcomb}$ 
            & Number of convex combinations for each new sample. 
            \\
        \end{tabular}
        \end{flushleft}
    }
    \SetKwInOut{Parameter}{Parameters}
    \Parameter{
        \begin{flushleft}
        \begin{tabular}{ l l l }
            $\var{neb\_conv}$ 
            & $(\geq 1)$
            & Number of data points used in affine combination for new samples. 
            \\
            $\var{shadow}$ 
            & $(\geq 1)$
            & Number of generated shadowsamples per parent data point. \\
            $\var{sigma}$ 
            & $(\geq 0)$
            & List of standard deviations for normal distributions for adding noise to each feature.
            \\
        \end{tabular}
        \end{flushleft}
    }
    
    
    \vspace{1em}

    \textbf{Function} \var{generate\_points}(\var{cluster}, \var{num\_samples}, \var{num\_convcomb}) \Begin{
    \SetKwFor{For}{For}{do}{endfor}
    \SetKwIF{If}{ElseIf}{Else}{If}{then}{else If}{else}{endif}

    Initialize \var{generated\_data} with empty set.
    \vspace{5pt}

    \eIf{$|\var{cluster}| > \var{neb\_conv}$}{
        $\var{neb\_list} \leftarrow{} \mbox{set of all $k$-neighbourhoods in \var{cluster}}$
    }{
        $\var{neb\_list} \leftarrow{} \{ \var{cluster} \}$
    }
    \vspace{3pt}

    \eIf{$\var{num\_convcomb} < \var{num\_feats}$}{
        $k \leftarrow{} 2$
    }{
        $k \leftarrow{} \var{num\_convcomb}$
    }
    \vspace{3pt}

    \For{$i = 1, 2, \ldots$ \var{num\_samples}}{
        $\var{neighbourhood} \leftarrow{} $ a random neighbourhood in \var{neb\_list}
        \vspace{3pt}

        \eIf{$\var{num\_convcomb} < \var{num\_feats}$}{
            $\var{data\_shadow} \leftarrow{} \var{neighbourhood}$
        }{
            Initialize \var{data\_shadow} with empty set.

            \For{$v \in neighboururhood$}{
                $\var{data\_shadow} \leftarrow \var{data\_shadow} \cup \{ \var{shadow}$ random vectors around $v$ with normal distribution. $\}$
            }
        }
        \vspace{3pt}

        $u = (u_1, \ldots, u_k) \leftarrow{} k$ random vectors $\in \var{data\_shadow}$

        $w = (w_1, \ldots, w_k) \leftarrow{} $ a random vector with positive values and $w_1 + w_2 + \ldots + w_k=1$
        \vspace{3pt}

        $\var{generated\_data} \leftarrow{} \var{generated\_data} \cup \{ \sum_{i = 1}^k w_i \cdot u_i \}$
    }
    Returns new points as \var{generated\_data}.
    }
\end{algorithm*}

The motivation behind the ProWRAS algorithm arises from a pilot study we performed on $14$ imbalanced datasets using four different classifiers GB, RF, kNN, and LR. We observed that using the oversampling algorithms described in Section \ref{Algo} are variously efficient on different classifiers. We discuss the results and their implications of our pilot study in detail in Section \ref{sec: results}. The ProWRAS algorithm is a multi-schematic oversampling algorithm, which integrates several aspects of the LoRAS and ProWSyn algorithm. The ProWRAS algorithm can be realised by the following steps:

\textbf{Partition/Cluster minority class samples:} The algorithm takes labelled imbalanced data as input. As a first step, it creates a partition of the minority class. The partition is done as per the proximity of the minority class data points from the majority class. The maximum number of desired partitions can be predefined by the user using a parameter $\var{max\_levels}$ (recommended value of $5$). The first partition $P_1$ is determined by the union of $\var{n\_neighbours\_max}$ (recommended value of $5$) number of minority class nearest neighbours of all the majority class data points. The parameter $\var{n\_neighbours\_max}$ can also be adjusted by the user. Once the first partition $P_1$ is ready, the process is repeated for the remaining minority class data points (if any left) that are not in $P_1$. This procedure is repeated for $L-1$ steps to obtain partitions $P_1, \dots, P_{L-1}$. The minority class data points that are not included in any of the partitions  $P_1, \dots, P_{L-1}$, form the partition $P_L$. Thus, for $i<j$, $P_i$ is closer to the majority class compared to $P_j$, $i,j \in \{1,\dots, L\}$. The partitions thus formed are treated as clusters in the minority class. This clustering technique is adopted from ProWSyn, a very effective oversampling technique, described in Section \ref{Algo} \cite{ProWSyn}. An advantage of this type of partitioning/ clustering of data is that the clustering process considers the distribution of the minority class with respect to the majority class, which is not considered in other clustering algorithms.

\textbf{Assigning proximity weights to clusters:} The next step is to assign proximity weights to each cluster $P_i$, $i \in \{1,\dots, L\}$, such that clusters closer to the majority class have more weights. This is done to make the decision boundary stronger than, the minority class data points that are closer to the majority class cause more confusion for the classifiers to create a decision boundary. This is achieved for $i \in \{1,\dots, L\}$, by assigning weight $w_i$ to the cluster $P_i$, following the equation:
\begin{equation}\label{weights}
    w_i=e^{-\theta\cdot(i-1)}
\end{equation}
The weights are then normalised. The parameter $\theta$ (recommended value $1$) can be used to control the rate of decay of weights. The pseudocode for this process can be found in Algorithm \ref{algoProWras:partitionInfo}.

\textbf{Deciding the number of synthetic samples to generate from each cluster:}
After the clusters and their respective normalized weights are obtained, ProWRAS decides the number of synthetic samples to be generated from each cluster. The total number of synthetic samples to be generated is taken as a user input using the parameter $\var{num\_samples\_to\_generate}$ (a recommended value for this parameter is the difference between the number of majority and minority class samples). The normalised weights of respective clusters are multiplied to $\var{num\_samples\_to\_generate}$, to determine the number of samples to be generated from those clusters. Until this point, the ProWRAS algorithm follows the same steps as the ProWSyn algorithm \cite{ProWSyn}.

\textbf{Customize variance for each cluster:}
Bej \textit{et al.}\ in the article on the LoRAS algorithm pointed out that customising the variance of the synthetic samples can be important for an improved modelling of the convex space of the minority class. In contrast to the ProWSyn algorithm, the ProWRAS algorithm uses an approach to rigorously model the convex space of the minority class by controlling the variance of the synthetic samples generated. There are two aspects of the algorithm that help us achieve this, which we describe below.
\begin{figure*}[!htbp]
\caption{Illustration of the working principle of the ProWRAS algorithm. ProWRAS used a proximity based partitioning system to find clusters in the minority class. For each cluster, it then uses one of four oversampling schemes shown in the figure. The key to success of the ProWRAS algorithm is its ability to rigorously model the convex space through controlling the variance of the synthetic samples.}
\label{ProWRAS_figure_main}
\centering
\includegraphics[scale=.85]{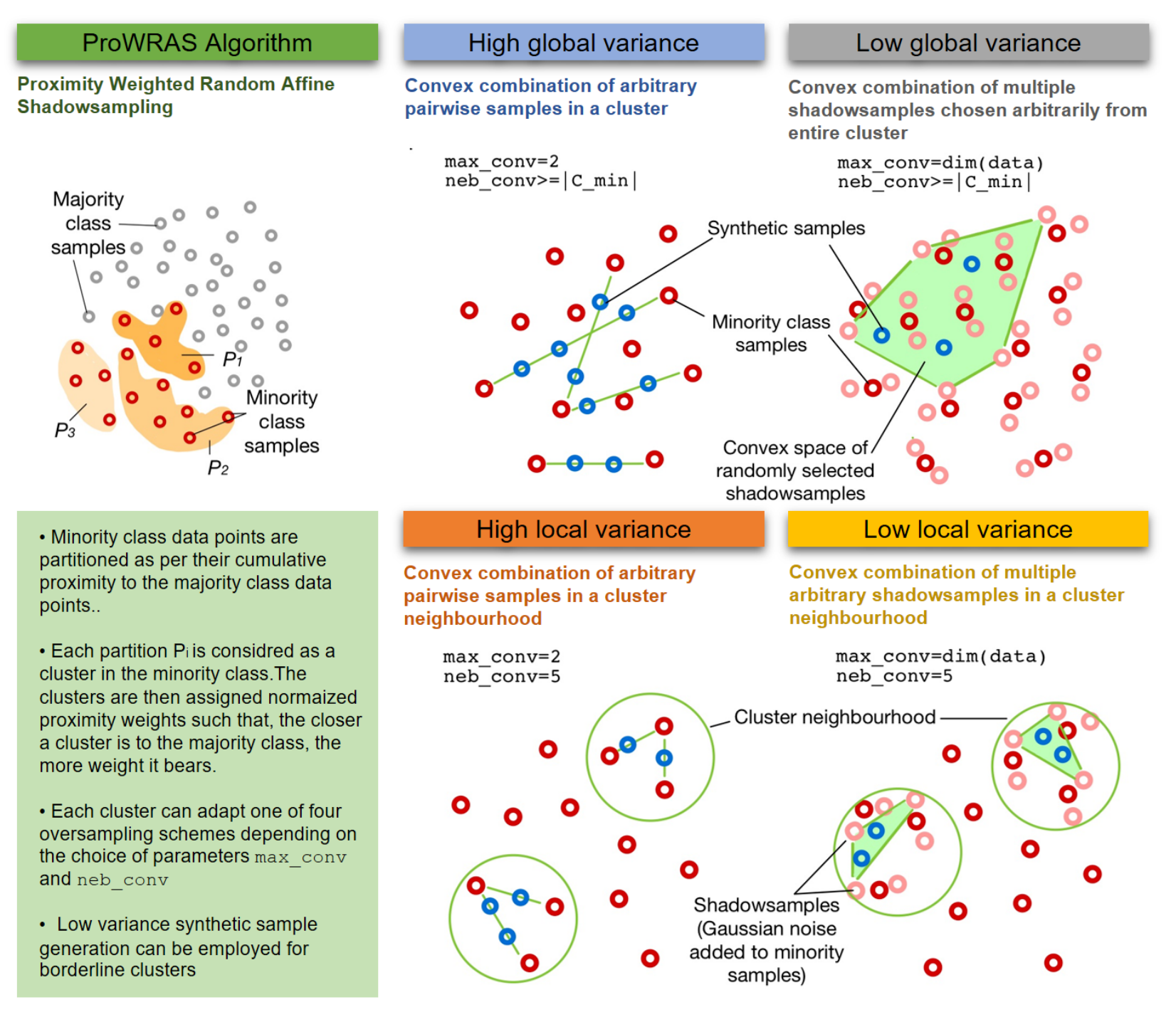}  
\end{figure*}

\textbf{A) Choice of proper neighbourhood size:} For each of the identified clusters, ProWRAS can generate synthetic samples using different minority class neighbourhood sizes. This can lead to a local oversample generation scheme by choosing a small minority class neighbourhood (similar to the approach of SMOTE or LoRAS) or a global oversample generation scheme, which considers the entire cluster as a neighbourhood (similar to the approach of CURE-SMOTE). The global or local oversampling schemes can be accessed using proper choice of $\var{neb\_conv}$ parameter. If the choice of $\var{neb\_conv}$ is less than the size of the cluster itself, then ProWRAS will employ local oversampling scheme, otherwise the whole cluster will be considered as a neighbourhood and the global oversampling scheme is employed. 
    
\textbf{B) Convex space modelling:} ProWRAS can also control the variance of the generated synthetic samples using rigorous convex space modelling. This is achieved using the $\var{max\_conv}$ parameter. If $\var{max\_conv}=2$, ProWRAS generates SMOTE-like synthetic samples by taking convex combinations of any two minority class samples. We call this, a high variance oversampling scheme, since this leads to high variance of the synthetic samples (see Equation \ref{LoRASvar}). If $\var{max\_conv}>2$, ProWRAS generates LoRAS-like synthetic samples by taking convex combinations of multiple numbers of shadowsamples. LoRAS-like sample generation of course requires two more parameters to be added to the algorithm: $\sigma$ (recommended value of $0.001$), for deciding the variance of the normal distribution to draw the noise for creating the shadowsamples, $\var{shadow}$ (recommended value of $100$), for deciding how many shadowsamples need to be created per minority class sample. We call this a low variance oversampling scheme. The number of convex combinations is decided by the normalized proximity weight of the respective cluster. We first scale the normalized proximity weights of all clusters, dividing them by the maximum normalized proximity weight obtained. Note that, this produces scaled weights for every cluster, such that each cluster has a weight between $0$ and $1$. The scaled weights are then multiplied with the  $\var{max\_conv}$ parameter to obtain the number of appropriate convex combinations of shadowsamples for each cluster.

Note that, clusters with higher scaled weights are closer to the majority class. Taking a convex combination of multiple samples for such clusters will help to keep the variance of the synthetic samples low, which will prevent them from interfering with the majority class. Clusters with lower scaled weights are far away from the majority class, and hence we can choose to create high variance synthetic samples from them. This also reduces the computational costs of the LoRAS algorithm.

\begin{figure*}[!htbp]
\scriptsize
\caption{Illustration of the study design for our experiments.}
\label{study design}
\centering
\includegraphics[scale=1.1]{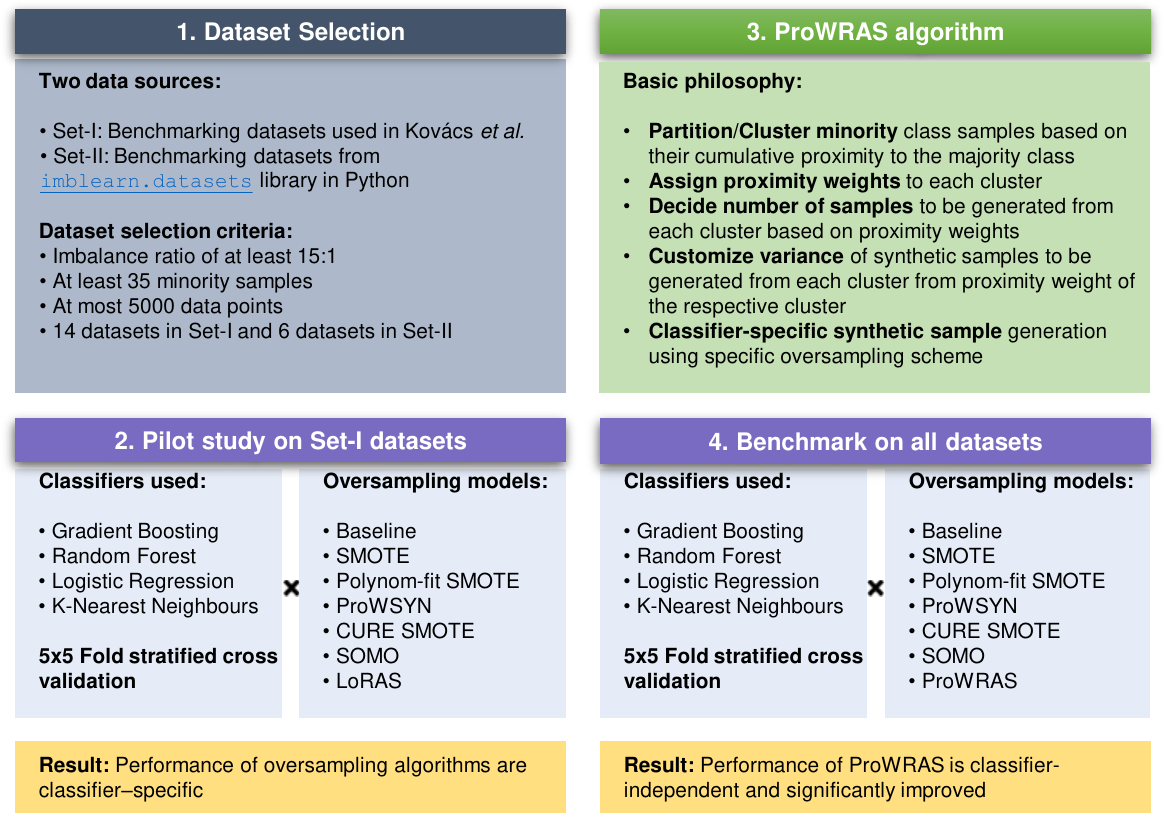}  
\end{figure*}

\begin{table*}[!htbp]
\scriptsize
\caption{Table showing the ProWRAS oversampling scheme used for every dataset and for every classifier. HGV: High global variance, LGV: Low global variance, HLV: High local variance, LLV: Low local variance. Column 2-5 show the oversampling scheme for which ProWRAS works best for respective datasets and classifiers. Furthermore, the table shows some statistics for the datasets. Datasets from Set II are marked in red.}
\label{Schemesandstats}
\centering
\tabularnewline
\begin{tabular}{l |@{\hskip3pt}c@{\hskip3pt}|@{\hskip3pt} c @{\hskip3pt}|@{\hskip3pt} c@{\hskip3pt}|@{\hskip3pt} c@{\hskip3pt}|@{\hskip3pt} c@{\hskip3pt}|@{\hskip3pt} c@{\hskip3pt}|@{\hskip3pt} c@{\hskip3pt}}
\hline
Dataset & GB & RF & kNN & LR & Imbalance ratio & Minority samples & Total samples \\ [0.5ex] 
\hline\hline
abalone9-18 & HGV & HGV & LGV & HGV & 16.40 & 42 & 731\\
\hline
abalone\_17\_vs\_7\_8\_9\_10 & HGV & HGV & LGV & LLV & 39.31 & 58 & 2338
\\
\hline
car-vgood & LLV & HLV & LGV & LGV  & 25.58 & 65 & 1728 \\
\hline
car\_good & LGV & HLV & LLV & LGV & 24.04 & 69 & 1728 \\
\hline
flare-F & HGV & HGV & LLV & LGV & 23.79 & 43 & 1066 \\
\hline
hypothyroid & LGV & LGV & LLV & LGV  & 19.95 & 151  & 3163\\
\hline
kddcup-guess\_passwd\_vs\_satan &  HLV &  HGV &  LLV &  LGV & 29.98 & 53 & 1642\\
\hline
kr-vs-k-three\_vs\_eleven & LGV & HGV & LGV & LGV  & 35.23 & 81 & 2935\\
\hline
kr-vs-k-zero-one\_vs\_draw & LGV & LGV & LGV & LGV & 26.63 & 105 & 2901\\
\hline
shuttle-2\_vs\_5 &  HGV &  HGV &  LLV &  LGV & 66.67 & 49 & 3316\\
\hline
winequality-red-4 & HLV & HGV & HLV & LGV  & 29.17 & 53 & 1599\\
\hline
yeast4 & HLV & HGV & LGV & LGV  & 28.10 & 51 & 1484\\
\hline
yeast5 & HLV & HGV & LGV & LLV & 32.73 & 44 & 1484\\
\hline
yeast6 & LGV & HLV & LGV & LLV  & 41.40 & 35 & 1484\\
\hline
 \textcolor{red}{oil} & HGV & HGV & LGV & LGV  & 22.85 & 41 & 937\\
\hline
\textcolor{red}{ozone\_level} & HGV & HGV & LLV & LGV & 34.73 & 73 & 2536\\
\hline
\textcolor{red}{solar\_flare\_m0} & HLV & HLV & LLV & LGV  & 20.42 & 68 & 1389\\
\hline
\textcolor{red}{thyroid\_sick} & LGV & HLV & LLV & LGV & 16.32 & 231 & 3772\\
\hline
\textcolor{red}{wine\_quality} & HLV & HLV & HLV & LGV & 26.76 & 183 & 4898\\
\hline
\textcolor{red}{yeast\_me2} & LLV & HLV & LLV & LGV & 29.09 & 51 & 1484\\
\hline

\end{tabular}
\end{table*}

Based on the points mentioned above, we can identify four oversampling schemes for the ProWRAS algorithm that can be accessed by different combinations of the two parameters $\var{max\_conv}$ and $\var{neb\_conv}$. They are:
\begin{itemize}
    \item High global variance (HGV)\\ ($\var{max\_conv}=2$, $\var{neb\_conv}\geq|C_\text{min}|$)
    \item Low global variance (LGV)\\
    ($\var{max\_conv}=\var{dim(data)}$, $\var{neb\_conv}\geq|C_\text{min}|$)
    \item High local variance (HLV)\\
    ($\var{max\_conv}=2$, $\var{neb\_conv}=5$)
    \item Low local variance (LLV)\\
    ($\var{max\_conv}=\var{dim(data)}$, $\var{neb\_conv}=5$)
\end{itemize}
where $C_\text{min}$ is the minority class and  $\var{dim(data)}$ is the number of features in the dataset. Note that the global oversampling scheme is employed for a cluster, if the chosen value of $\var{neb\_conv}$ is greater than the size of the cluster. Choosing the value of $\var{neb\_conv} \geq |C_\text{min}|$ ensures that for all clusters the global oversampling scheme is employed. Moreover, if $\var{max\_conv}=2$ then automatically scales cluster weights to determine the number of convex combinations for the shadowsample generation in any arbitrary cluster becomes $2$ (See Algorithm \ref{algoProWras:generatePoints}), leading ProWRAS into the high variance data generation scheme. A graphical representation of the ProWRAS algorithm is shown in Figure \ref{ProWRAS_figure_main}.
To sum up, the ProWRAS algorithm has eight adjustable parameters: $\var{max\_levels}$, $\var{n\_neighbours\_max}$, $\var{num\_samples\_to\_generate}$, $\theta$, $\sigma$, $\var{shadow}$, $\var{max\_conv}$, $\var{neb\_conv}$. Among these, only two parameters $\var{max\_conv}$, $\var{neb\_conv}$, affect the oversampling process significantly by changing the use of one of the four oversampling schemes.\\
\textbf{Classifier-specific synthetic sampling:} Finally, to take full advantage of the ProWRAS algorithm given a dataset and a classifier of choice, a user can choose to train the classifier using all four oversampling schemes and finally select the best one. In Section \ref{sec: results}, where we discuss our results, we show that, classifier-specific choice of oversampling schemes helps ProWRAS to perform better, independently of the classifier used.

\section{Case studies}\label{sec:case studies}

\subsection{Benchmarking datasets}\label{Benchmarking datasets}
For our pilot study and our final benchmarking study, we selected two sets of public datasets. The first set (Set-I) is a subset of the $104$ publicly available imbalanced datasets used for the benchmarking studies in Kovács \textit{et al.}\ \cite{Comparison}. The second set (Set-II), is a subset of $27$ publicly available datasets in the \href{https://imbalanced-learn.org/stable/generated/imblearn.datasets.fetch_datasets.html}{{\fontfamily{pcr}\selectfont imblearn.datasets}} Python library. Set-I has $14$ datasets and Set-II has $6$ datasets. We select the datasets, Set-I and Set-II based on the following three criteria. Note that, we select all the datasets following all the three criteria, to ensure that our choice of datasets for the studies are impartial. The criteria are:  
\begin{itemize}
    \item We choose datasets with an imbalance ratio of at least $15:1$. This is to ensure that the performance of the compared oversampling algorithms are tested, particularly on datasets with high imbalance.\\
    \item We choose datasets with a minimum of $35$ minority class samples. Datasets with very less minority class samples, classifier performances that are often affected by high stochasticity and the results are often statistically unreliable. Setting this condition for the choice of datasets, enhances the reliability of results.\\
    \item We choose datasets with at most $5000$ samples (only $4$ datasets do not satisfy this condition). Given that our studies involve $4$ classifiers, $8$-different oversampling models and $4$-oversampling schemes of the ProWRAS algorithm, we consider this constraint to limit the computational effort.
\end{itemize}

In Table \ref{Schemesandstats} we show the statistics for the relevant datasets. 
\subsection{Protocols for benchmarking}\label{Protocols for benchmarking}
The datasets of Set-I are used for the pilot study, comparing Baseline classification with oversampling models SMOTE, Polynom-fit SMOTE, ProWSyn, CURE-SMOTE, SOMO, and LoRAS for four respective classifiers GB, RF, kNN, and LR. For our final benchmarking study, we use all 20 datasets from Set I and Set II, comparing Baseline classification with oversampling models SMOTE, Polynom-fit SMOTE, ProWSyn, CURE-SMOTE, LoRAS, and ProWRAS for four classifiers GB, RF, kNN, and LR. For both our pilot study and for our main benchmarking study, we use $5\times5$ stratified cross validation as validation protocol. While training our models, we use oversampling algorithms only on the training data for each fold. Also, we use normalised datasets for training and testing.

\textbf{Choice of classification models and parameters:}\\
We have used the classification models GB, RF, kNN, and LR for our benchmarking studies. We use GB and RF, since they are powerful classifiers that use ensemble approaches of boosting and bagging, respectively. kNN and LR were also seen to perform well for imbalanced datasets in the benchmarking study of Bej \textit{et al.}\ \cite{LoRAS}. Both in our pilot study and in the final benchmarking study, we used default parameters for all the classifiers as recommended in {\fontfamily{pcr}\selectfont scikit-learn (V 0.21.2)} documentation.

\textbf{Choice of oversampling models and parameters:}\\
In our final benchmarking study, we compare five benchmarking algorithms against ProWRAS. We choose these algorithms in particular for the following reasons:
\begin{itemize}
    \item SMOTE, of course, is the pioneer of all algorithms and still, widely used because of its simplicity and applicability.\\
    \item Polynom-fit SMOTE, ProWSyn are the top two oversampling algorithms by overall performance, from the detailed benchmarking study by Kov\'acs \cite{Comparison}.\\
    \item ProWSyn, CURE-SMOTE, and SOMO also use the idea of using clustering approaches on the minority class to learn the distribution of the minority class better and take advantage of it during synthetic sample generation, a philosophy they share with ProWRAS algorithm. Moreover, both CURE-SMOTE and SOMO are proposed fairly recently(in $2017$). SOMO has been used only in the pilot study. It was excluded in the main study because of its low $\mathscr{I}$-score in the pilot experiment (See Table \ref{pilot_CI}).\\
    \item We chose LoRAS because, evidently, ProWRAS is an extension of the LoRAS algorithm.
\end{itemize}

\begin{table*}[h!]
\scriptsize
\caption{Table showing F1-score/$\kappa$-score for several oversampling strategies (Baseline, SMOTE, Polynom-fit SMOTE, ProWSyn, CURE-SMOTE, LoRAS, ProWRAS) for all 20 benchmarking datasets for Gradient Boosting classifier. The column on the right shows the performance of the ProWRAS algorithm over all datasets. We observe in the last row that the average performance of ProWRAS is superior to all the other oversampling algorithms.}
\label{table_imbsk_GB}
\centering
\tabularnewline
\begin{tabular}{l |@{\hskip3pt}c@{\hskip3pt}|@{\hskip3pt} c @{\hskip3pt}|@{\hskip3pt} c@{\hskip3pt}|@{\hskip3pt} c@{\hskip3pt}| @{\hskip3pt}c @{\hskip3pt}|@{\hskip3pt}c@{\hskip3pt}|@{\hskip3pt} c@{\hskip3pt}}
\hline
Dataset & Baseline & SMOTE & Polynom-fit SMOTE & ProWSyn & CURE-SMOTE & LoRAS & \textbf{ProWRAS}\\ [0.5ex] 
\hline\hline
abalone9-18 & 0.342/0.319 & 0.359/0.312 & 0.259/0.236 & 0.381/0.338 & 0.317/0.271 & 0.319/0.294 & 0.385/0.341 \\
\hline
abalone\_17\_vs\_7\_8\_9\_10 & 0.277/0.265 & 0.333/0.308 & 0.294/0.282 & 0.359/0.336 & 0.337/0.314 & 0.236/0.226 & 0.335/0.310 
\\
\hline
car-vgood & 0.981/0.980 & 0.946/0.943 & 0.966/0.964 & 0.968/0.967 & 0.960/0.959 & 0.968/0.966 & 0.959/0.957 \\
\hline
car\_good & 0.900/0.896 & 0.850/0.843 & 0.819/0.812 & 0.855/0.849 & 0.84/0.833 & 0.868/0.863 & 0.863/0.857 \\
\hline
flare-F & 0.172/0.155 & 0.321/0.287 & 0.174/0.156 & 0.271/0.241 & 0.143/0.123 & 0.171/0.149 & 0.320/0.291 \\
\hline
hypothyroid & 0.805/0.796 & 0.762/0.748 & 0.798/0.788 & 0.776/0.763 & 0.788/0.778 & 0.796/0.787 & 0.803/0.794 \\
\hline
kddcup-guess\_passwd\_vs\_satan &  1.0/1.0 &  1.0/1.0 &  1.0/1.0 &  1.0/1.0 &  1.0/1.0 & 1.0/1.0 & 0.998/0.998 \\
\hline
kr-vs-k-three\_vs\_eleven & 0.993/0.992 & 0.993/0.993 & 0.995/0.995 & 0.993/0.993 & 0.995/0.995 & 0.995/0.995 & 0.995/0.995 \\
\hline
kr-vs-k-zero-one\_vs\_draw & 0.969/0.968 & 0.949/0.947 & 0.969/0.968 & 0.958/0.957 & 0.962/0.96 & 0.961/0.96 & 0.969/0.967 \\
\hline
shuttle-2\_vs\_5 &  1.0/1.0 &  1.0/1.0 &  1.0/1.0 &  1.0/1.0 &  1.0/1.0 &  1.0/1.0 &  1.0/1.0 \\
\hline
winequality-red-4 & 0.056/0.045 & 0.154/0.112 & 0.083/0.044 & 0.150/0.105 & 0.132/0.093 & 0.143/0.105 & 0.156/0.115 \\
\hline
yeast4 & 0.304/0.287 & 0.351/0.32 & 0.307/0.290 & 0.339/0.305 & 0.321/0.299 & 0.214/0.197 & 0.335/0.303 \\
\hline
yeast5 & 0.697/0.689 & 0.739/0.730 & 0.727/0.719 & 0.734/0.725 & 0.738/0.731 & 0.703/0.695 & 0.735/0.726 \\
\hline
yeast6 & 0.454/0.444 & 0.456/0.44 & 0.505/0.496 & 0.462/0.445 & 0.507/0.496 & 0.554/0.545 & 0.514/0.501 \\
\hline
 oil & 0.482/0.464 & 0.525/0.503 & 0.482/0.466 & 0.537/0.517 & 0.527/0.507 & 0.549/0.533 & 0.587/0.568 \\
\hline
ozone\_level & 0.166/0.154 & 0.341/0.317 & 0.221/0.207 & 0.332/0.306 & 0.233/0.221 & 0.269/0.254 & 0.329/0.303 \\
\hline
solar\_flare\_m0 & 0.106/0.085 & 0.155/0.115 & 0.110/0.088 & 0.129/0.109 & 0.121/0.101 & 0.098/0.078 & 0.194/0.139 \\
\hline
thyroid\_sick & 0.865/0.856 & 0.852/0.841 & 0.858/0.849 & 0.845/0.834 & 0.845/0.836 & 0.856/0.847 & 0.873/0.864 \\
\hline
wine\_quality & 0.232/0.216 & 0.278/0.234 & 0.22/0.178 & 0.247/0.201 & 0.225/0.189 & 0.238/0.193 & 0.290/0.248 \\
\hline
yeast\_me2 & 0.329/0.313 & 0.361/0.33 & 0.319/0.303 & 0.339/0.305 & 0.207/0.192 & 0.329/0.241 & 0.225/0.328 \\
\hline
\textbf{Average}     & 0.556/0.546 & 0.586/0.566 & 0.555/0.542 & 0.584/0.565 & 0.564/0.549 & 0.560/0.546 & 0.600/0.580 \\
\hline

\end{tabular}
\end{table*}

\begin{table*}[h!]
\scriptsize
\caption{Table showing F1-score/$\kappa$- score for several oversampling strategies (Baseline, SMOTE, Polynom-fit SMOTE, ProWSyn, CURE-SMOTE, LoRAS, and ProWRAS) for all 20 benchmarking datasets for Random Forest classifier. The column on the right shows the performance of the ProWRAS algorithm over all datasets. We observe in the last row that the average performance of ProWRAS is superior to all the other oversampling algorithms.}
\label{table_imbsk_RF}
\centering
\tabularnewline
\begin{tabular}{l |@{\hskip3pt}c@{\hskip3pt}|@{\hskip3pt} c @{\hskip3pt}|@{\hskip3pt} c@{\hskip3pt}|@{\hskip3pt} c@{\hskip3pt}| @{\hskip3pt}c @{\hskip3pt}|@{\hskip3pt}c@{\hskip3pt}|@{\hskip3pt} c@{\hskip3pt}}
\hline
Dataset & Baseline & SMOTE & Polynom-fit SMOTE & ProWSyn & CURE-SMOTE & LoRAS & \textbf{ProWRAS}\\ [0.5ex] 
\hline\hline
abalone9-18 & 0.211/0.198 & 0.342/0.3 & 0.266/0.245 & 0.327/0.278 & 0.311/0.272 & 0.325/0.293 & 0.38/0.335 \\
\hline
abalone\_17\_vs\_7\_8\_9\_10 & 0.17/0.166 & 0.338/0.318 & 0.25/0.241 & 0.324/0.3 & 0.242/0.228 & 0.247/0.233 & 0.328/0.303 \\
\hline
car-vgood & 0.959/0.958 & 0.974/0.973 & 0.937/0.935 & 0.955/0.953 & 0.922/0.919 & 0.943/0.942 & 0.972/0.971 \\
\hline
car\_good & 0.78/0.773 & 0.8/0.793 & 0.713/0.705 & 0.768/0.76 & 0.723/0.715 & 0.6/0.59 & 0.869/0.863 \\
\hline
flare-F & 0.087/0.066 & 0.147/0.117 & 0.091/0.07 & 0.183/0.156 & 0.075/0.055 & 0.1/0.08 & 0.21/0.181 \\
\hline
hypothyroid & 0.786/0.777 & 0.755/0.742 & 0.791/0.782 & 0.756/0.743 & 0.783/0.773 & 0.785/0.775 & 0.787/0.778 \\
\hline
kddcup-guess\_passwd\_vs\_satan & 0.998/0.998 & 0.998/0.998 &  1.0/1.0 & 0.998/0.998 & 0.998/0.998 & 0.998/0.998 & 0.998/0.998 \\
\hline
kr-vs-k-three\_vs\_eleven & 0.989/0.989 & 0.995/0.995 & 0.989/0.989 & 0.991/0.991 & 0.991/0.991 & 0.991/0.990 & 0.996/0.996 \\
\hline
kr-vs-k-zero-one\_vs\_draw & 0.961/0.96 & 0.949/0.947 & 0.955/0.953 & 0.958/0.956 & 0.956/0.954 & 0.941/0.939 & 0.961/0.959 \\
\hline
shuttle-2\_vs\_5 &  1.0/1.0 &  1.0/1.0 &  1.0/1.0 &  1.0/1.0 &  1.0/1.0 &  1.0/1.0 &  1.0/1.0 \\
\hline
winequality-red-4 & 0.007/0.007 & 0.113/0.086 & 0.048/0.025 & 0.168/0.131 & 0.037/0.022 & 0.104/0.078 & 0.158/0.119 \\
\hline
yeast4 & 0.238/0.23 & 0.351/0.326 & 0.26/0.249 & 0.352/0.321 & 0.267/0.251 & 0.245/0.235 & 0.357/0.325 \\
\hline
yeast5 & 0.655/0.647 & 0.751/0.743 & 0.723/0.716 & 0.739/0.73 & 0.695/0.687 & 0.723/0.716 & 0.754/0.746 \\
\hline
yeast6 & 0.435/0.428 & 0.478/0.465 & 0.464/0.456 & 0.474/0.459 & 0.482/0.473 & 0.531/0.523 & 0.518/0.507 \\
\hline
oil & 0.391/0.379 & 0.52/0.503 & 0.429/0.415 & 0.586/0.57 & 0.441/0.427 & 0.421/0.408 & 0.58/0.564 \\
\hline
ozone\_level & 0.014/0.012 & 0.301/0.284 & 0.086/0.081 & 0.326/0.304 & 0.114/0.109 & 0.177/0.17 & 0.33/0.307 \\
\hline
solar\_flare\_m0 & 0.07/0.047 & 0.105/0.066 & 0.058/0.036 & 0.109/0.079 & 0.067/0.045 & 0.081/0.058 & 0.168/0.117 \\
\hline
thyroid\_sick & 0.843/0.833 & 0.875/0.867 & 0.839/0.829 & 0.85/0.84 & 0.843/0.834 & 0.85/0.841 & 0.868/0.859 \\
\hline
wine\_quality & 0.292/0.282 & 0.378/0.355 & 0.304/0.283 & 0.331/0.296 & 0.275/0.258 & 0.307/0.282 & 0.368/0.344 \\
\hline
 yeast\_me2 & 0.181/0.173 & 0.406/0.384 & 0.223/0.211 & 0.359/0.328 & 0.242/0.229 & 0.268/0.257 & 0.389/0.365 \\
\hline
\textbf{Average} & 0.504/0.496 & 0.579/0.563 & 0.521/0.511 & 0.578/0.56 & 0.524/0.519 & 0.531/0.521 & 0.600/0.582 \\
\hline

\end{tabular}
\end{table*}

\begin{table*}[h!]
\scriptsize
\caption{Table showing F1-score/$\kappa$- score for several oversampling strategies (Baseline, SMOTE, Polynom-fit SMOTE, ProWSyn, CURE-SMOTE, LoRAS, and ProWRAS) for all 20 benchmarking datasets for k-Nearest neighbours classifier. The column on the right shows the performance of the ProWRAS algorithm over all datasets. We observe in the last row that the average performance of ProWRAS is superior to all the other oversampling algorithms.}
\label{table_imbsk_KNN}
\centering
\tabularnewline
\begin{tabular}{l |@{\hskip3pt}c@{\hskip3pt}|@{\hskip3pt} c @{\hskip3pt}|@{\hskip3pt} c@{\hskip3pt}|@{\hskip3pt} c@{\hskip3pt}| @{\hskip3pt}c @{\hskip3pt}|@{\hskip3pt}c@{\hskip3pt}|@{\hskip3pt} c@{\hskip3pt}}
\hline
Dataset & Baseline & SMOTE & Polynom-fit SMOTE & ProWSyn & CURE-SMOTE & LoRAS & \textbf{ProWRAS}\\ [0.5ex] 
\hline\hline
abalone9-18 & 0.206/0.197 & 0.285/0.223 & 0.345/0.295 & 0.316/0.257 & 0.335/0.279 & 0.309/0.249 & 0.384/0.35 \\
\hline
abalone\_17\_vs\_7\_8\_9\_10 & 0.104/0.1 & 0.276/0.247 & 0.339/0.316 & 0.324/0.297 & 0.295/0.269 & 0.297/0.271 & 0.347/0.328 \\
\hline
car-vgood & 0.827/0.822 & 0.74/0.728 & 0.737/0.723 & 0.703/0.688 & 0.765/0.753 & 0.741/0.728 & 0.818/0.81 \\
\hline
car\_good & 0.581/0.571 & 0.645/0.625 & 0.607/0.585 & 0.491/0.458 & 0.645/0.626 & 0.645/0.626 & 0.641/0.622 \\
\hline
flare-F & 0.203/0.183 & 0.273/0.228 & 0.257/0.212 & 0.255/0.208 & 0.284/0.244 & 0.311/0.273 & 0.301/0.263 \\
\hline
hypothyroid & 0.437/0.422 & 0.479/0.444 & 0.509/0.48 & 0.487/0.454 & 0.546/0.523 & 0.545/0.519 & 0.553/0.528 \\
\hline
kddcup-guess\_passwd\_vs\_satan &  1.0/1.0 & 0.996/0.996 & 0.994/0.994 &  1.0/1.0 &  1.0/1.0 &  0.994/0.994 &  1.0/1.0 \\
\hline
kr-vs-k-three\_vs\_eleven & 0.902/0.899 & 0.918/0.915 & 0.925/0.923 & 0.911/0.908 & 0.907/0.904 & 0.92/0.917 & 0.94/0.938 \\
\hline
kr-vs-k-zero-one\_vs\_draw & 0.852/0.847 & 0.867/0.861 & 0.879/0.874 & 0.872/0.866 & 0.868/0.863 & 0.875/0.87 & 0.9/0.896 \\
\hline
shuttle-2\_vs\_5 &  1.0/1.0 &  1.0/1.0 &  1.0/1.0 &  1.0/1.0 &  1.0/1.0 &  1.0/1.0 &  1.0/1.0 \\
\hline
winequality-red-4 & 0.039/0.036 & 0.132/0.083 & 0.13/0.081 & 0.13/0.079 & 0.145/0.099 & 0.13/0.08 & 0.141/0.093 \\
\hline
yeast4 & 0.219/0.21 & 0.261/0.217 & 0.286/0.244 & 0.267/0.223 & 0.292/0.25 & 0.257/0.212 & 0.308/0.268 \\
\hline
yeast5 & 0.69/0.682 & 0.631/0.616 & 0.655/0.642 & 0.614/0.598 & 0.677/0.665 & 0.661/0.648 & 0.714/0.704 \\
\hline
yeast6 & 0.574/0.565 & 0.314/0.288 & 0.363/0.339 & 0.301/0.274 & 0.507/0.493 & 0.4/0.379 & 0.588/0.579 \\
\hline
oil & 0.329/0.319 & 0.426/0.39 & 0.475/0.444 & 0.456/0.423 & 0.476/0.445 & 0.492/0.464 & 0.545/0.531 \\
\hline
ozone\_level & 0.165/0.155 & 0.202/0.161 & 0.202/0.162 & 0.2/0.159 & 0.23/0.192 & 0.216/0.177 & 0.218/0.179 \\
\hline
solar\_flare\_m0 & 0.052/0.033 & 0.226/0.168 & 0.207/0.146 & 0.2/0.138 & 0.208/0.155 & 0.225/0.175 & 0.228/0.177 \\
\hline
thyroid\_sick & 0.5/0.48 & 0.527/0.487 & 0.531/0.495 & 0.528/0.49 & 0.555/0.525 & 0.556/0.524 & 0.556/0.526 \\
\hline
wine\_quality & 0.104/0.096 & 0.246/0.2 & 0.226/0.177 & 0.22/0.168 & 0.235/0.19 & 0.232/0.185 & 0.25/0.204 \\
\hline
yeast\_me2 & 0.255/0.245 & 0.309/0.27 & 0.287/0.248 & 0.266/0.223 & 0.306/0.272 & 0.329/0.295 & 0.339/0.305 \\
\hline
\textbf{Average} & 0.452/0.443 & 0.488/0.457 & 0.498/0.469 & 0.477/0.446 & 0.514/0.487 & 0.506/0.479 & 0.538/0.515 \\

\hline

\end{tabular}
\end{table*}

\begin{table*}[h!]
\scriptsize
\caption{Table showing F1-score/$\kappa$- score for several oversampling strategies (Baseline, SMOTE, Polynom-fit SMOTE, ProWSyn, CURE-SMOTE, LoRAS, and ProWRAS) for all 20 benchmarking datasets for Logistic Regression classifier. The column on the right shows the performance of the ProWRAS algorithm over all datasets. We observe in the last row that the average performance of ProWRAS is superior to all the other oversampling algorithms.}
\label{table_imbsk_LR}
\centering
\tabularnewline
\begin{tabular}{l |@{\hskip3pt}c@{\hskip3pt}|@{\hskip3pt} c @{\hskip3pt}|@{\hskip3pt} c@{\hskip3pt}|@{\hskip3pt} c@{\hskip3pt}| @{\hskip3pt}c @{\hskip3pt}|@{\hskip3pt}c@{\hskip3pt}|@{\hskip3pt} c@{\hskip3pt}}
\hline
Dataset & Baseline & SMOTE & Polynom-fit SMOTE & ProWSyn & CURE-SMOTE & LoRAS & \textbf{ProWRAS}\\ [0.5ex] 
\hline\hline
abalone9-18 & 0.463/0.448 & 0.46/0.413 & 0.488/0.447 & 0.481/0.437 & 0.447/0.401 & 0.464/0.418 & 0.488/0.446 \\
\hline
abalone\_17\_vs\_7\_8\_9\_10 & 0.266/0.259 & 0.301/0.271 & 0.333/0.307 & 0.3/0.271 & 0.309/0.281 & 0.312/0.284 & 0.315/0.287 \\
\hline
car-vgood & 0.086/0.075 & 0.378/0.338 & 0.403/0.365 & 0.388/0.35 & 0.37/0.33 & 0.382/0.343 & 0.407/0.371 \\
\hline
car\_good &  0.0/0.0 & 0.099/0.028 & 0.1/0.029 & 0.099/0.028 & 0.102/0.033 &  0.096/0.025 & 0.103/0.033 \\
\hline
flare-F & 0.213/0.198 & 0.268/0.217 & 0.313/0.268 & 0.272/0.222 & 0.298/0.253 & 0.268/0.218 & 0.341/0.3 \\
\hline
hypothyroid & 0.356/0.339 & 0.386/0.338 & 0.42/0.376 & 0.39/0.342 & 0.423/0.381 & 0.412/0.367 & 0.446/0.407 \\
\hline
kddcup-guess\_passwd\_vs\_satan & 0.99/0.99 & 0.99/0.99 & 0.99/0.99 & 0.99/0.99 & 0.99/0.99 & 0.99/0.99 & 0.99/0.99 \\
\hline
kr-vs-k-three\_vs\_eleven & 0.921/0.919 & 0.877/0.873 & 0.924/0.922 & 0.898/0.895 & 0.864/0.86 & 0.881/0.878 & 0.928/0.926 \\
\hline
kr-vs-k-zero-one\_vs\_draw & 0.816/0.81 & 0.692/0.677 & 0.796/0.787 & 0.727/0.714 & 0.716/0.703 & 0.713/0.699 & 0.853/0.847 \\
\hline
shuttle-2\_vs\_5 & 0.966/0.966 &  1.0/1.0 &  1.0/1.0 &  1.0/1.0 &  1.0/1.0 & 1.0/1.0 &  1.0/1.0 \\
\hline
winequality-red-4 & 0.007/0.007 & 0.113/0.086 & 0.048/0.025 & 0.168/0.131 & 0.037/0.022 & 0.104/0.078 & 0.158/0.119 \\
\hline
yeast4 & 0.219/0.21 & 0.261/0.217 & 0.286/0.244 & 0.267/0.223 & 0.292/0.25 & 0.257/0.212 & 0.308/0.268 \\
\hline
yeast5 & 0.511/0.5 & 0.571/0.552 & 0.615/0.599 & 0.584/0.566 & 0.601/0.584 & 0.585/0.567 & 0.591/0.574 \\
\hline
yeast6 & 0.383/0.375 & 0.309/0.281 & 0.33/0.304 & 0.298/0.269 & 0.388/0.366 & 0.316/0.289 & 0.326/0.3 \\
\hline
oil & 0.524/0.507 & 0.474/0.44 & 0.518/0.49 & 0.506/0.475 & 0.484/0.454 & 0.475/0.444 & 0.535/0.511 \\
\hline
ozone\_level & 0.191/0.178 & 0.262/0.226 & 0.287/0.253 & 0.263/0.226 & 0.314/0.282 & 0.311/0.279 & 0.347/0.319 \\
\hline
solar\_flare\_m0 & 0.112/0.102 & 0.196/0.124 & 0.216/0.151 & 0.204/0.133 & 0.206/0.142 & 0.192/0.119 & 0.25/0.195 \\
\hline
thyroid\_sick & 0.644/0.626 & 0.511/0.465 & 0.652/0.624 & 0.534/0.492 & 0.611/0.579 & 0.535/0.492 & 0.57/0.532 \\
\hline
wine\_quality & 0.083/0.077 & 0.178/0.121 & 0.195/0.14 & 0.188/0.133 & 0.225/0.175 & 0.169/0.111 & 0.196/0.143 \\
\hline
yeast\_me2 & 0.22/0.21 & 0.255/0.211 & 0.28/0.238 & 0.262/0.218 & 0.291/0.25 & 0.266/0.222 & 0.295/0.254 \\
\hline
\textbf{Average} & 0.399/0.39 & 0.429/0.393 & 0.46/0.428 & 0.441/0.406 & 0.448/0.417 & 0.436/0.401 & 0.472/0.441 \\

\hline

\end{tabular}
\end{table*}

For the pilot study, we used default parameters for SMOTE, Polynom-fit SMOTE, ProWSyn, CURE-SMOTE, and SOMO algorithms. For LoRAS we use parameter values of $\var{k}=5$, $\var{|S\_p|}=100$, $\var{L\textsubscript{\textsigma}}=5\times 10^{-8}$ and regular  $\var{embedding}$. For the parameter $\var{L\textsubscript{\textsigma}}$, we use a random search among the values $\{2,10,30,\var{dim(data)})\}$. This random search is based on a training and testing (disjoint sets) done on randomly chosen $50$ percent and $20$ percent subset of the respective dataset, chosen such that the imbalance ratio is maintained in the randomly chosen subsets. 

For our final benchmarking study, we also used default parameters for SMOTE, Polynom-fit SMOTE, ProWSyn, CURE-SMOTE, and LoRAS algorithms. However, the oversampling neighbourhood for all the oversampling algorithms (wherever applicable) is assumed to be $5$, observing that the minority class for all the chosen datasets is rather small. For ProWRAS we use fixed parameter values of $\var{max\_levels}=5$, $\var{n\_neighbours\_max}=5$, $\var{num\_samples\_to\_generate}=|C_{\text{max}}|-|C_{\text{min}}|$, $\theta=1$, $\var{shadow}=100$ and $\sigma = 10^{-6}$. These choices are the recommended default values of the algorithm and have been decided upon by performing independent pilot studies on datasets not used in the benchmarking experiments (e.g. \href{https://www.kaggle.com/mlg-ulb/creditcardfraud}{credit fraud dataset}). To access the four oversampling schemes, we use four combinations of values for the parameters $\var{max\_conv}$, $\var{neb\_conv}$.
\begin{itemize}
    \item High global variance (HGV)\\ ($\var{max\_conv}=2$, $\var{neb\_conv}=1000$)
    \item Low global variance (LGV)\\
    ($\var{max\_conv}=\var{dim(data)}$, $\var{neb\_conv}=1000$)
    \item High local variance (HLV)\\
    ($\var{max\_conv}=2$, $\var{neb\_conv}=5$)
    \item Low local variance (LLV)\\
    ($\var{max\_conv}=\var{dim(data)}$, $\var{neb\_conv}=5$)
\end{itemize}
The choice of proper oversampling scheme is based on a training and testing (disjoint sets) done on randomly chosen $50$ percent and $20$ percent subset of the respective dataset, chosen such that the imbalance ratio is maintained in the randomly chosen subsets. In Table \ref{Schemesandstats}, we present, for every classifier and every dataset, which oversampling scheme worked best among the four. Note that, we previously discussed that, for accessing the global oversampling scheme for a certain cluster, $\var{neb\_conv}$ must be at least the size of the cluster. Since all our chosen datasets have minority class size of less than $1000$, a choice of $\var{neb\_conv}=1000$ to access the global oversampling schemes works for all datasets.

\textbf{Performance measures:}
Choice of performance measures are an important aspect of studies with imbalanced datasets. For our study, we used two performance measures, F1-score and Cohen's $\kappa$-score. F1-score is the harmonic mean of precision and recall and is a good measure for how good the classification is for the minority class. Given a classification problem, the $\kappa$ measure is formally defined as:
\begin{equation*}
    \kappa=\frac{P_o-P_e}{1-P_e}
\end{equation*}
where, $P_o$ is the measure of observed agreement among the chosen classifier and the ground truths of the classification problem. $P_e$ is the measure of agreement by chance, among a chosen classifier and the ground truths of the classification problem. probability of agreement. The $\kappa$-score gives us a quantification of how good the classification is, considering both the majority and the minority class. 

For coding, we used the {\fontfamily{pcr}\selectfont scikit-learn (V 0.21.2)}, {\fontfamily{pcr}\selectfont  numpy (V 1.16.4)}, {\fontfamily{pcr}\selectfont  pandas (V 0.24.2)}, and {\fontfamily{pcr}\selectfont  matplotlib (V 3.1.0)} libraries in {\fontfamily{pcr}\selectfont  Python (V 3.7.4)}.

We provide an implementation of our algorithm and several Jupyter notebooks from the benchmarking study in  \href{https://github.com/COSPOV/ProWRAS} {GitHub}.

\section{Results}\label{sec: results}
\begin{table*}[ht!]
\scriptsize
\caption{Table showing the $\mathscr{I}$-scores for different oversampling algorithms for the pilot study.}
\label{pilot_CI}
\centering
\tabularnewline
\begin{tabular}{l |@{\hskip3pt}c@{\hskip3pt}|@{\hskip3pt} c @{\hskip3pt}|@{\hskip3pt} c@{\hskip3pt}|@{\hskip3pt} c@{\hskip3pt}| @{\hskip3pt}c @{\hskip3pt}|@{\hskip3pt}c@{\hskip3pt}|@{\hskip3pt} c@{\hskip3pt}}
\hline
  & Baseline & SMOTE & Polynom-fit SMOTE & ProWSyn & CURE-SMOTE & SOMO & LoRAS\\ [0.5ex] 
\hline\hline
$\mathscr{I}$ & 0.442 & 0.58 & 0.592 & 0.562 & 0.649 & 0.478 & 0.647 \\
\hline

\end{tabular}
\end{table*}

\begin{table*}[ht!]
\scriptsize
\caption{Table showing the $\mathscr{I}$-scores for different oversampling algorithms for the final benchmarking experiments.}
\label{main_CI}
\centering
\tabularnewline
\begin{tabular}{l |@{\hskip3pt}c@{\hskip3pt}|@{\hskip3pt} c @{\hskip3pt}|@{\hskip3pt} c@{\hskip3pt}|@{\hskip3pt} c@{\hskip3pt}| @{\hskip3pt}c @{\hskip3pt}|@{\hskip3pt}c@{\hskip3pt}|@{\hskip3pt} c@{\hskip3pt}}
\hline
 & Baseline & SMOTE & Polynom-fit SMOTE & ProWSyn & CURE-SMOTE & LoRAS & ProWRAS\\ [0.5ex] 
\hline\hline
$\mathscr{I}$ & 0.301 & 0.484 & 0.403 & 0.505 & 0.524 & 0.496 & 0.833 \\
\hline

\end{tabular}
\end{table*}

\begin{figure*}[h!]
\caption{Figure showing results for the final study. Every heatmap for the respective classifier shows the number of datasets for which the oversampling model in the $i$-th row performs equally or better (by F1-score) than the model in the $j$-th column. Note that, ProWRAS performs consistently well for all the classifiers.} 
\label{final_study}
\centering
\includegraphics[scale=1]{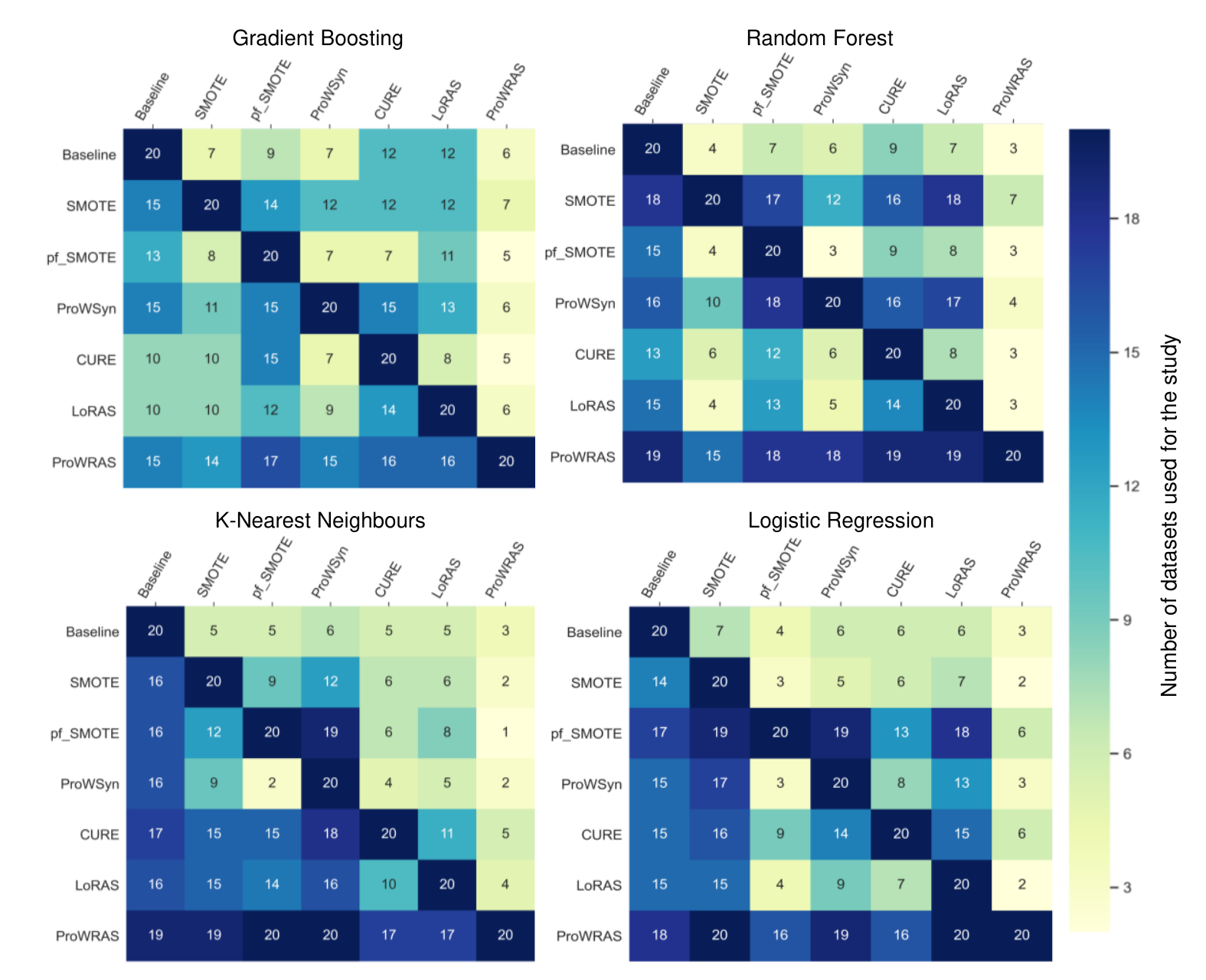}  
\end{figure*}

\textbf{Results for pilot study:} The detailed results for the pilot studies can be found in Supplementary data (Section \ref{suppl}). We observe from our pilot study, that for kNN classifier LoRAS, CURE-SMOTE, and Polynom-fit SMOTE are the best performers. For LR classifier, CURE-SMOTE and Polynom-fit SMOTE are ahead of the other oversampling models. For the RF classifier, SMOTE and ProWSyn are the best performers. For GB classifier, ProWSyn, LoRAS, and SMOTE generate better F1-scores.

Moreover, we also observe that the average F1-score and $\kappa$-score for all classifiers is comparatively better for the ensemble based classifiers RF and GB (see Supplementary data in Section \ref{suppl}).

We also provide the $\mathscr{I}$-score for every oversampling model used in the pilot study in Table \ref{pilot_CI}. We observe that CURE-SMOTE, Polynom-fit SMOTE and LoRAS produce the best scores, while the score of SOMO is quite close to the baseline. Thus, we excluded the SOMO algorithm in our final benchmarking studies.

\textbf{Results for final benchmarking study:}
Detailed results of our final benchmarking studies are shown in Table \ref{table_imbsk_GB},\ref{table_imbsk_RF},\ref{table_imbsk_KNN},\ref{table_imbsk_LR}. We observe that the ensemble models on an average perform quite well in comparison to kNN or LR. Even the baseline model for GB has better F1 and $\kappa$-scores compared to the oversampled classifiers for kNN and LR. For classifiers GB, RF, and LR, ProWRAS produces better average F1-score and $\kappa$-score over all models. In Figure \ref{final_study} we show heatmap plots for each chosen classifier, showing comparisons of the oversampling algorithms among each other in terms of their performance on the $20$. We observe that the oversampling models that performed well for the pilot study for respective classifiers continue to do well. For example, for RF, SMOTE and ProWSYN still perform quite well and for kNN, LoRAS and CURE-SMOTE still performs consistently. ProWRAS performs consistently well for all the classifiers. 

 We have also quantified the classifier independence of the compared oversampling algorithms using the $\mathscr{I}$-score (\ref{CI}). We observed that, CURE-SMOTE, LoRAS and ProWSYN still maintains comparatively high degree of classifier independence. However, ProwRAS significantly outperforms other algorithms with a score of $0.833$. 
Thus, we conclude that, classifier and dataset specific synthetic data generation of ProWRAS makes its performance classifier-independent.

\begin{table*}[!htbp]
\scriptsize
\caption{Table showing the results of Wilcoxon's signed rank test for comparison of ProWRAS with other oversampling algorithms. The p-values in the table quantify the statistical significance of the improvement achieved by ProWRAS over each oversampling model. Higher value of $W_{+}$, shows that how superior performance of ProWRAS for higher number of datasets. Higher value of $R$ shows better reliability of the results.}
\label{Wilcox}
\centering
\tabularnewline
\begin{tabular}{l |@{\hskip3pt}c@{\hskip3pt}|@{\hskip3pt} c @{\hskip3pt}|@{\hskip3pt} c@{\hskip3pt}|@{\hskip3pt} c@{\hskip3pt}| @{\hskip3pt}c @{\hskip3pt}|@{\hskip3pt}c@{\hskip3pt}|@{\hskip3pt} c@{\hskip3pt}}
\hline
Dataset & classifier & Baseline & SMOTE & Polynom-fit SMOTE & ProWSYN & CURE-SMOTE & LoRAS \\ [0.5ex] 
\hline\hline

p-value (F1 score) & GB & 0.003 & 0.024 & 0.000 & 0.012 & 0.001 & 0.003  \\
\hline
p-value ($\kappa$-score) &	GB &	0.006 &	0.017 &	0.000 &	0.014 &	0.001 &	0.003 \\
\hline
$W_+$/$W_-$ (F1 score) &	GB &	189/15 &	182/21 &	195/6 &	189/15 & 195/10 &	189/10  \\
\hline
$W_+$/$W_-$ ($\kappa$-score) &	GB &	189/15 &	189/15 & 195/6 &	189/15 &	195/10 &	189/10 	\\
\hline
$R$ (F1 score) &	GB & 0.860 &	0.795 &	0.925 &	0.860 &	0.914 &	0.878  \\
\hline
$R$ ($\kappa$-score) &	GB &	0.860 &	0.860 &	0.925 &	0.860 &	0.914 &	0.878 \\
\hline
\hline

p-value (F1 score) & RF & 0.000 & 0.013 & 0.000 & 0.001 & 0.000 & 0.000  \\
\hline
p-value ($\kappa$-score) &	RF &	0.000 &	0.019 &	0.000 &	0.001 &	0.000 &	0.000 \\
\hline
$W_+$/$W_-$ (F1 score) &	RF &	204/1 &	182/15 &	204/3 &	200/3 &	207/0 &	207/1  \\
\hline
$W_+$/$W_-$ ($\kappa$-score) &	RF &	204/1 &	174/21 & 204/3 &	200/3 &	207/0 &	207/1 	\\
\hline
$R$ (F1 score) &	RF &	0.983 &	0.820 &	0.989 &	0.960 &	0.995 &	1  \\
\hline
$R$ ($\kappa$-score) &	RF &	0.983 &	0.750 &	0.989 &	0.960 &	0.995 &	1 \\
\hline
\hline

p-value (F1 score) & kNN & 0.000 & 0.000 & 0.000 & 0.000 & 0.001 & 0.000  \\
\hline
p-value ($\kappa$-score) &	kNN  & 0.000 & 0.000 & 0.000 & 0.000 & 0.001 & 0.000 \\
\hline
$W_+$/$W_-$ (F1 score) &	kNN &	204/1 &	207/1 &	209/0 &	207/0 &	195/6 &	200/6  \\
\hline
$W_+$/$W_-$ ($\kappa$-score) &	kNN &	204/1 &	207/1 &	209/0 &	207/0 &	195/6 &	204/3	\\
\hline
$R$ (F1 score) &	kNN &	0.983 &	1 &	1 &	0.995 &	0.925 &	0.957  \\
\hline
$R$ ($\kappa$-score) &	kNN &	0.983 &	1 &	1 &	0.995 &	0.925 &	0.984 \\
\hline
\hline

p-value (F1 score) & LR & 0.001 & 0.000 & 0.047 & 0.000 & 0.034 & 0.003  \\
\hline
p-value ($\kappa$-score) &	LR &	0.007 &	0.000 &	0.047 &	0.000 &	0.034 &	0.000 \\
\hline
$W_+$/$W_-$ (F1 score) &	LR &	204/3 &	207/0 &	189/10 &	204/1 &	189/10 &	207/0  \\
\hline
$W_+$/$W_-$ ($\kappa$-score) &	LR &	200/6 &	207/0 & 182/15 &	204/1 &	189/10 &	207/ 10 	\\
\hline
$R$ (F1 score) &	LR &	0.989 &	0.995 &	0.878 &	0.983 &	0.878 &	0.995  \\
\hline
$R$ ($\kappa$-score) &	LR &	0.957 &	0.995 &	0.820 &	0.983 &	0.878 &	0.995 \\
\hline

\end{tabular}
\end{table*}
\begin{figure*}[h!]
\caption{Distribution of the oversampling schemes used by the ProWRAS algorithm and by each classifier over all investigated 20 benchmarking datasets. For GB, we see that all four oversampling schemes has similar frequencies over the datasets. For RF, we observe that high variance schemes are preferred over low variance schemes. For KNN and LR, we observe that LLV and LGV are most effective, respectively.}
\label{scheme_stats}
\centering
\includegraphics[scale=.7]{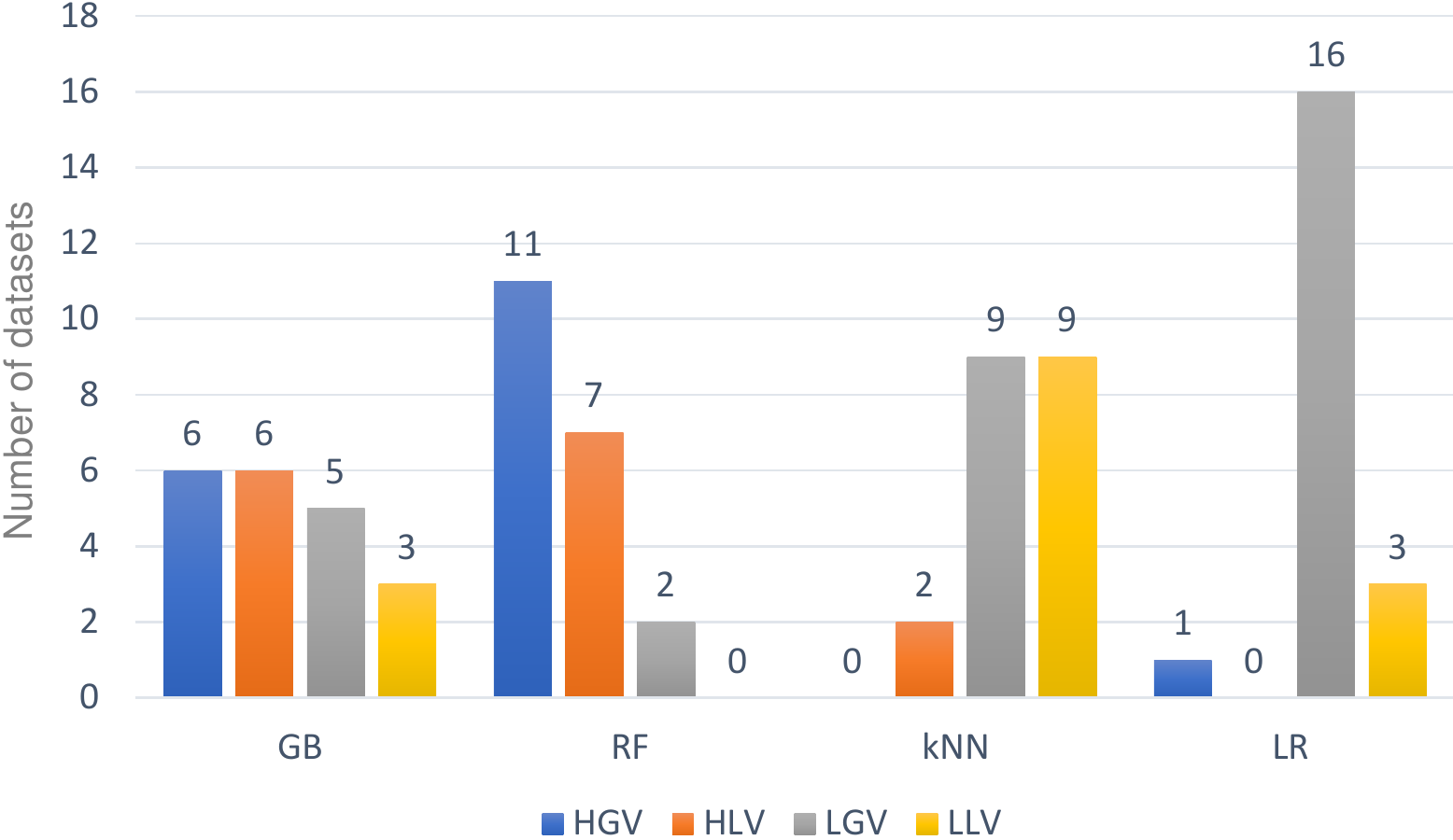}  
\end{figure*} 
\textbf{Statistical significance of results:}
A statistical significance test quantifies how significant the improvement of a model is, compared to another arbitrary model. Here, we use the Wilcoxon's Signed Rank Test (WSRT) that was already used in related recent articles on imbalanced datasets \cite{LoRAS, SMOTEFUNA}. The null hypothesis $H_0$ for WSRT, is that, there is not a significant difference between the performance of two compared algorithms. A p-value obtained after exercising the test, if below the threshold of $0.05$, then the null hypothesis can be rejected. However, only the p-value is not informative enough \cite{SMOTEFUNA}. Hence, three more metrics $W_+$, $W_-$, and $R$ are used (see \cite{LoRAS, SMOTEFUNA}). 

To make the article self-contained, we borrow the descriptions of these metrics from Bej \textit{et al.}\cite{LoRAS}:

\begin{itemize}
    \item ``For each data pair (involving LoRAS and some other oversampling algorithm) of model predictions, the difference between both predictions is calculated and stored in a vector $D$, excluding the zero difference values. 
    \item The signs of the difference is recorded in a sign vector $S$. 
    \item The entries in $|D|$ are ranked, forming a vector $R'$. In case of tied ranks, an average ranking scheme is adopted. This means, after ranking the entries of $|D|$ are ranked using integers and then, in case of tied entries, the average of the integer ranks are considered as the average rank for all the respective tied entries with a specific tied value.
    \item Component-wise product of $S$ and $R'$ provides us with the vector $W$, the vector of the signed ranks. The sum of absolute values of the positive entries in $W$ is $W_+$ and the  sum of absolute values of the negative entries in $W$ is $W_-$. After this we define, $W_R=min\{W_+,W_-\}$
    \item Then the test statistic $Z$ is calculated by the equation
    \begin{equation}
    \begin{split}
    Z= \frac{W_R-\frac{n(n+1)}{4}}{\sqrt{\frac{n(n+1)(2n+1)}{24}-\frac{\Sigma t^3-\Sigma t}{48}}}
    \end{split}
    \end{equation}
    where $n$ is the number of components in $D$ and $t$ is the number of times some $i$-th entry occurs in $R'$, summed over all such repeated instances.
    \item Finally, $R$ is calculated using $R=\frac{|Z|}{\sqrt{N}}$, where $N$ is the total number of datasets compared, which is $14$ in our case. \cite{LoRAS} ''
\end{itemize}

The higher the value of, $W_+$ the better the performance of ProWRAS with respect to the compared algorithms, whereas a higher value of $W_-$ implies the opposite. The value of $R$ quantifies the degree of improvement of ProWRAS compared to the other oversampling models. In Table \ref{Wilcox} we provide the p-values, as well as the $W_+$, $W_-$ and $R$ measures for ProWRAS against other oversampling models for all the classifiers used in our study. 

From Table \ref{Wilcox}, we can observe that ProWRAS significantly improves classifier performance compared to other oversampling algorithms in most cases. A few exceptions are SMOTE for RF and GB, Polynom-fit SMOTE and CURE-SMOTE for LR etc. But even in these cases from Figure \ref{final_study}, we can see that for both GB and RF, SMOTE does equally or better than ProWRAS in $7$ out of $20$ datasets, for LR, Polynom-fit SMOTE does equally or better than ProWRAS in only $3$ out of $20$ datasets and CURE-SMOTE does better than ProWRAS in $6$ out of $20$ datasets. We can thus conclude, that even though ProWRAS does not improve model performance significantly in these few cases, it certainly performs at par with the compared model for these cases.

\section{Discussion}\label{sec: discussion}

There are of course certain limitations of the ProWRAS oversampling approach. For instance, oversampling models based on modelling the convex space of the minority class are more effective for datasets where the convex space of the data can add some meaningful information to the training experience of the classifiers. ProWRAS also relies on convex space modelling of the minority class and hence can be widely applicable to regular tabular data that are homogeneous in nature, that is the features are well-defined with respect to the classification problem. Moreover, the approach is also more suitable for datasets with more feature variables are continuous in nature rather than ordinal or nominal feature variables.

From Figure \ref{scheme_stats}, we observe that the distribution of oversampling schemes used by ProWRAS used by LoRAS is vastly different for different classifiers. For KNN, the LGV and LLV oversampling scheme has been largely effective, for LR, the LGB scheme has worked out for most datasets. For RF, the oversampling schemes HGV and HLV have worked out to be the best. For GB, which has produced the best average F1-score and $\kappa$-score among all classifiers over all oversampling models, interestingly, have a more uniform distribution for the ProWRAS oversampling schemes.

For RF and GB classifiers except for ProWRAS, SMOTE, and ProWSyn also performs well. Note that SMOTE uses synthetic samples with high local variance (SMOTE generate synthetic samples from neighbourhoods of minority class data points and does not control the variance of the synthetic samples, hence high local variance), while ProWSyn, within each cluster adopts the high global variance philosophy of synthetic sample generation (ProWSyn generate synthetic samples from entire clusters of minority class data points and does not control the variance the synthetic samples, hence high global variance). Note that, for ProWRAS we also observe that for most datasets, HGV and HLV strategies are successful for both GB and RF. For LR except for ProWRAS, Polynom fit-SMOTE also performs well.

For the default star topology, Polynom fit-SMOTE controls the variance of the synthetic samples by choosing convex combination of a minority samples with the centroid of the minority class, rather than choosing convex combination of two minority samples. Moreover, Polynom fit-SMOTE does not generate synthetic samples from individual data neighbourhoods. Thus, it follows a global oversampling scheme. Although the oversampling strategy of Polynom fit-SMOTE does not follow any of the strategies we considered for ProWRAS exactly, the use of the star topology (the default topology used by Polynom fit-SMOTE, hence used by us in benchmarking studies), is quite similar to the LGV strategy, which again is the successful strategy used by ProWRAS for most datasets for LR.

For the kNN model, the LoRAS algorithm that produces synthetic samples with low local variance has also proved to be successful for our pilot study. LoRAS uses LLV strategy LoRAS both reduces the variance of synthetic samples and also generate them from neighbourhoods of each minority class data point. We see that the ProWRAS algorithm is also quite successful for the kNN classifier, using the LLV scheme and `low variance' strategies in general. 

We observe that given a dataset and a classifier, it is hard to predict which oversampling scheme of ProWRAS would be most effective for that particular dataset and classifier. To obtain the best results, it is highly recommended using all four oversampling schemes and choose the scheme that is most effective. Since there are more than a hundred variants of SMOTE available and scores of ML based classifiers, it is difficult for a modeler to choose an appropriate classifier and oversampling model given a dataset. In this sense, ProWRAS provides a modeler with the advantage of performing benchmarking experiments on only four oversampling schemes, rather than using numerous oversampling algorithms from a pool of more than a hundred. Our experiments show that choosing appropriately from these four oversampling schemes consistently provides a superior classification performance.

In addition, from our results, we do observe some patterns on which oversampling schemes works better for which classifiers. For example, for GB and RF, oversampling schemes HGV and HLV are likely to be more effective; for kNN, the schemes LGV and LLV are likely to be more effective and for LR, the scheme LGV is likely to be more effective.

To sum up, the results of the benchmarking study quantitatively shows the ProWRAS improve classifier performances for all the chosen classifiers, with proper choice of oversampling scheme compared to the other state-of-the-art oversampling algorithms. Moreover, we quantify the significance of improvement induced by ProWRAS using the Wilcoxon's signed rank test, which proves the improvement induced by using ProWRAS is statistically significant.

\section{Conclusion}

Our study confirms that different classifiers adapt differently to the different approaches of oversampling algorithms to generate synthetic samples. The proposed ProWRAS approach can model the convex space of the minority class more rigorously than the LoRAS algorithm by controlling the variance of the synthetic samples better. ProWRAS achieves this through four unique oversampling schemes, as well as a proximity-weighted clustering system of the minority class data. The oversampling scheme of ProWRAS depends on two factors controlling the variance of the synthetic samples: neighbourhood size and convex space modelling. 

Moreover, ProWRAS allows generating low variance synthetic samples only in borderline clusters to avoid an overlap with the majority class, making the synthetic sample generation computationally cheaper compared to LoRAS. Using the multi-schematic approach of oversampling, ProWRAS significantly improves performance of classifiers in terms of both F1-score and $\kappa$-score compared to state-of-the-art oversampling models.

Through a novel performance measure $\mathscr{I}$-score, we have shown in this article that ProWRAS can be used in a more `classifier independent' way compared to other oversampling algorithms. This means that, with appropriate choice of oversampling scheme, ProWRAS customizes synthetic sample generation according to a classifier of choice and thereby reduce benchmarking efforts. Thus, ProWRAS is highly flexible to different classifiers and can find broad applicability in solving classification problems for real-world imbalanced datasets.



%


\ifCLASSOPTIONcompsoc
  \section*{Availability of code}
\else
  \section*{Availability of code}
\fi

To support transparency, re-usability, and reproducibility, we provide an implementation of the algorithm for binary classification problems using {\fontfamily{pcr}\selectfont  Python (V 3.7.4)} and several {\fontfamily{pcr}\selectfont  Jupyter Notebooks} from our benchmarking study in the GitHub repository: \url{https://github.com/COSPOV/ProWRAS}.

\ifCLASSOPTIONcompsoc
  \section*{Acknowledgments}
\else
  \section*{Acknowledgment}
\fi


We thank the German Network for Bioinformatics Infrastructure (de.NBI) and  Establishment of Systems Medicine Consortium in Germany e:Med for their support, as well as the German Federal Ministry for Education and Research (BMBF) program (FKZ 01ZX1709C) and the EU Social Fund (ESF/14-BM-A55-0027/18) for funding.

\ifCLASSOPTIONcaptionsoff
  \newpage
\fi



%

\clearpage
\newpage

\section{Supplementary data} \label{suppl}

\begin{figure*}[h!]
\caption{Figure showing results for the pilot study. Every heatmap for respective classifier shows the number of datasets for which the oversampling model in the $i$-th row performs equally or better (by F1-score) than the model in the $j$-th column. For example, for the gradient boosting classifier LoRAS performs equally or better than SMOTE for $10$ out of $14$ datasets. Note that, none of the oversampling models perform consistently well for all the classifiers.
}
\label{pilot_study}
\centering
\includegraphics[scale=1]{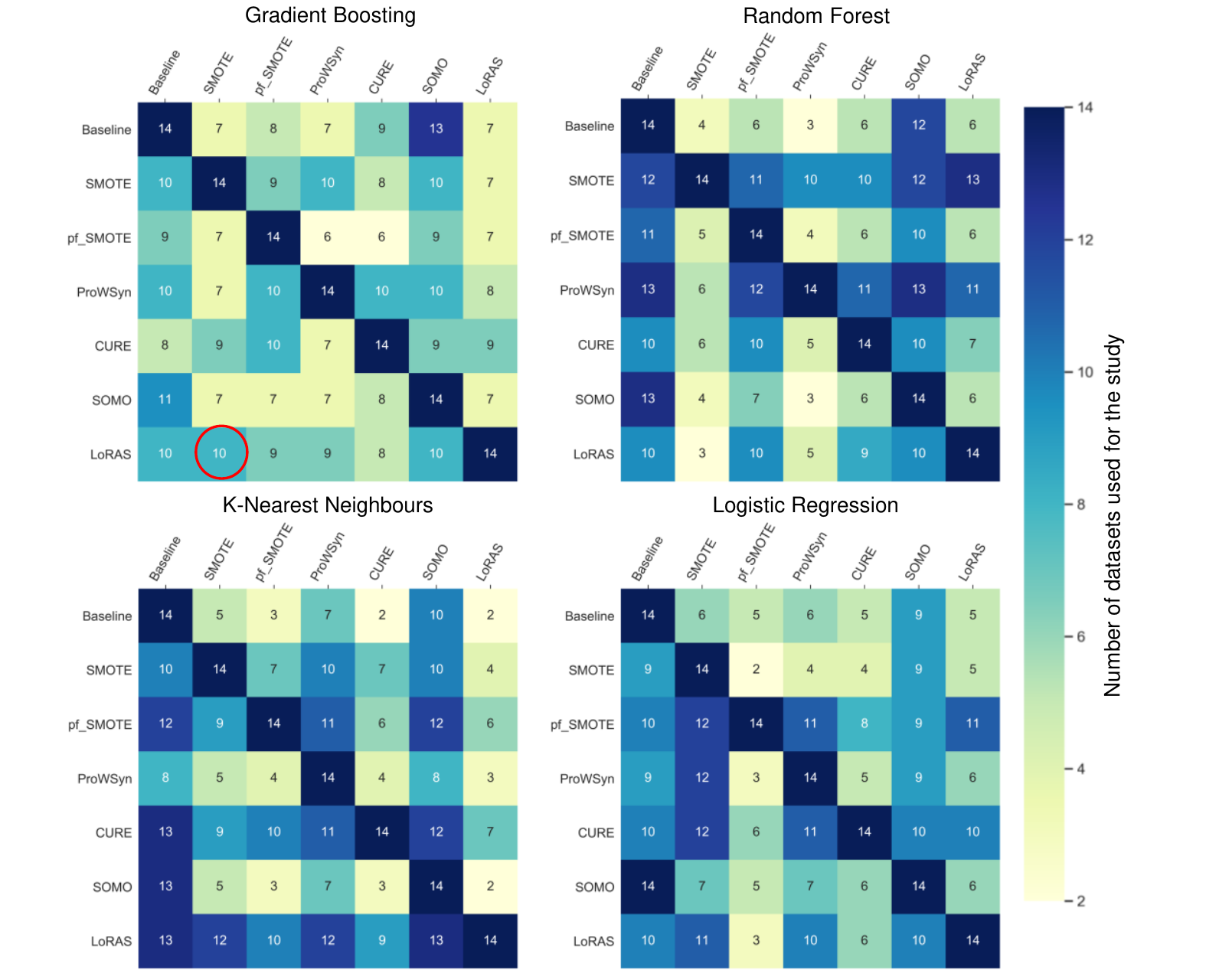}  
\end{figure*}

In Figure \ref{pilot_study}, we have shown a comparative plot for the performance by F1-score of all classifiers over all oversampling models.  Figure \ref{pilot_study} has a heatmap for every classifier. For a given classifier, the number in an arbitrary cell in the $i$-th row and $j$-th column, of the heatmap shows the number of datasets for which the oversampling model corresponding to the $i$-th row performs equally or better than the oversampling model corresponding to the $j$-th column. Since we have performed the pilot study on only the datasets of Set-I (with $14$ datasets), all diagonal elements are $14$. Notably, the performance by the $\kappa$ -score follows a  similar trend.

\begin{table*}[btp]
\tiny
\begin{minipage}[t][][b]{\textwidth}
\caption{Table showing F1-score/$\kappa$- score for several oversampling strategies (Baseline, SMOTE, Polynom-fit SMOTE, ProWSyn, CURE-SMOTE, SOMO, LoRAS) for $14$ Set-I benchmarking datasets for GB classifier.}
\label{table_pilot_GB}
\centering

\tabularnewline
\begin{tabular}{l |@{\hskip3pt}c@{\hskip3pt}|@{\hskip3pt} c @{\hskip3pt}|@{\hskip3pt} c@{\hskip3pt}|@{\hskip3pt} c@{\hskip3pt}| @{\hskip3pt}c @{\hskip3pt}|@{\hskip3pt}c@{\hskip3pt}|@{\hskip3pt}c@{\hskip3pt}}
\hline
     Datasets   &   Baseline &      SMOTE & Polynom-fit SMOTE &    ProWSyn &       CURE &       SOMO &     LORAS \\
\hline
\hline
abalone9-18 & 0.394/0.367 & 0.408/0.362 & 0.3/0.275 & 0.39/0.345 & 0.38/0.336 & 0.394/0.367 & 0.441/0.401 \\
\hline
abalone\_17\_vs\_7\_8\_9\_10 & 0.283/0.272 & 0.318/0.292 & 0.309/0.298 & 0.349/0.324 & 0.346/0.323 & 0.283/0.272 & 0.324/0.298 \\
\hline
car-vgood & 0.968/0.967 & 0.957/0.956 & 0.968/0.967 & 0.975/0.974 & 0.966/0.964 & 0.963/0.961 & 0.985/0.984 \\
\hline
car\_good & 0.904/0.9 & 0.813/0.805 & 0.86/0.855 & 0.839/0.832 & 0.867/0.862 & 0.893/0.888 & 0.867/0.862 \\
\hline
flare-F & 0.14/0.121 & 0.315/0.281 & 0.197/0.181 & 0.281/0.254 & 0.154/0.132 & 0.164/0.143 & 0.124/0.099 \\
\hline
hypothyroid & 0.806/0.797 & 0.781/0.769 & 0.81/0.801 & 0.776/0.763 & 0.792/0.782 & 0.806/0.797 & 0.793/0.782 \\
\hline
kddcup-guess\_passwd\_vs\_satan &  1.0/1.0 &  1.0/1.0 & 0.994/0.994 &  1.0/1.0 &  1.0/1.0 &  1.0/1.0 &  1.0/1.0 \\
\hline
kr-vs-k-three\_vs\_eleven &  1.0/1.0 &  1.0/1.0 &  1.0/1.0 &  1.0/1.0 &  1.0/1.0 &  1.0/1.0 &  1.0/1.0 \\
\hline
kr-vs-k-zero-one\_vs\_draw & 0.979/0.979 & 0.956/0.954 & 0.973/0.972 & 0.97/0.969 & 0.977/0.977 & 0.975/0.974 & 0.967/0.966 \\
\hline
shuttle-2\_vs\_5 &  1.0/1.0 &  1.0/1.0 &  1.0/1.0 &  1.0/1.0 &  1.0/1.0 &  1.0/1.0 &  1.0/1.0 \\
\hline
winequality-red-4 & 0.048/0.033 & 0.156/0.115 & 0.083/0.046 & 0.141/0.096 & 0.11/0.076 & 0.048/0.033 & 0.08/0.043 \\
\hline
yeast4 & 0.347/0.332 & 0.399/0.37 & 0.309/0.292 & 0.372/0.34 & 0.291/0.27 & 0.347/0.332 & 0.386/0.361 \\
\hline
yeast5 & 0.625/0.615 & 0.738/0.729 & 0.685/0.676 & 0.713/0.704 & 0.691/0.682 & 0.625/0.615 & 0.708/0.699 \\
\hline
yeast6 & 0.436/0.426 & 0.481/0.465 & 0.531/0.521 & 0.46/0.442 & 0.531/0.521 & 0.436/0.426 & 0.486/0.472 \\
\hline
Average & 0.638/0.629 & 0.666/0.65 & 0.644/0.634 & 0.662/0.646 & 0.65/0.637 & 0.638/0.629 & 0.654/0.641 \\

\hline

\end{tabular}

\end{minipage}

\begin{minipage}[t][][b]{\textwidth}
\caption{Table showing F1-score/$\kappa$- score for several oversampling strategies (Baseline, SMOTE, Polynom-fit SMOTE, ProWSyn, CURE-SMOTE, SOMO, LoRAS) for $14$ Set-I benchmarking datasets for RF classifier.}
\label{table_pilot_RF}
\centering
\tabularnewline
\begin{tabular}{l |@{\hskip3pt}c@{\hskip3pt}|@{\hskip3pt} c @{\hskip3pt}|@{\hskip3pt} c@{\hskip3pt}|@{\hskip3pt} c@{\hskip3pt}| @{\hskip3pt}c @{\hskip3pt}|@{\hskip3pt}c@{\hskip3pt}|@{\hskip3pt}c@{\hskip3pt}}
\hline
     Datasets   &   Baseline &      SMOTE & Polynom-fit SMOTE &    ProWSyn &       CURE &       SOMO &     LORAS \\
\hline
\hline
abalone9-18 & 0.28/0.449 & 0.389/0.347 & 0.313/0.29 & 0.336/0.287 & 0.336/0.298 & 0.28/0.267 & 0.368/0.331 \\
\hline
abalone\_17\_vs\_7\_8\_9\_10 & 0.106/0.353 & 0.339/0.319 & 0.167/0.157 & 0.32/0.296 & 0.278/0.261 & 0.106/0.103 & 0.31/0.29 \\
\hline
car-vgood & 0.969/0.995 & 0.988/0.988 & 0.954/0.952 & 0.974/0.973 & 0.947/0.945 & 0.969/0.967 & 0.932/0.93 \\
\hline
car\_good & 0.795/0.964 & 0.861/0.856 & 0.734/0.727 & 0.817/0.81 & 0.66/0.651 & 0.797/0.791 & 0.664/0.654 \\
\hline
flare-F & 0.08/0.232 & 0.182/0.151 & 0.025/0.003 & 0.196/0.169 & 0.045/0.024 & 0.027/0.004 & 0.086/0.06 \\
\hline
hypothyroid & 0.789/0.862 & 0.77/0.758 & 0.789/0.78 & 0.77/0.758 & 0.773/0.763 & 0.789/0.779 & 0.758/0.748 \\
\hline
kddcup-guess\_passwd\_vs\_satan &  1.0/1.0 &  1.0/1.0 &  1.0/1.0 &  1.0/1.0 &  1.0/1.0 &  1.0/1.0 &  1.0/1.0 \\
\hline
kr-vs-k-three\_vs\_eleven & 0.992/1.0 & 0.997/0.997 & 0.992/0.992 & 0.995/0.995 & 0.999/0.999 & 0.992/0.992 & 0.995/0.995 \\
\hline
kr-vs-k-zero-one\_vs\_draw & 0.95/0.997 & 0.943/0.941 & 0.95/0.948 & 0.955/0.953 & 0.951/0.949 & 0.953/0.951 & 0.939/0.937 \\
\hline
shuttle-2\_vs\_5 &  1.0/1.0 &  1.0/1.0 &  1.0/1.0 &  1.0/1.0 &  1.0/1.0 &  1.0/1.0 &  1.0/1.0 \\
\hline
winequality-red-4 & 0.0/0.172 & 0.143/0.115 & 0.031/0.01 & 0.164/0.127 & 0.054/0.037 & 0.0/-0.0 & 0.07/0.046 \\
\hline
yeast4 & 0.23/0.408 & 0.413/0.391 & 0.264/0.252 & 0.372/0.342 & 0.287/0.27 & 0.23/0.219 & 0.341/0.319 \\
\hline
yeast5 & 0.637/0.752 & 0.724/0.715 & 0.707/0.699 & 0.728/0.719 & 0.689/0.68 & 0.637/0.629 & 0.715/0.705 \\
\hline
yeast6 & 0.428/0.53 & 0.493/0.481 & 0.496/0.489 & 0.46/0.445 & 0.528/0.519 & 0.428/0.42 & 0.504/0.494 \\
\hline
Average & 0.59/0.694 & 0.66/0.647 & 0.602/0.593 & 0.649/0.634 & 0.61/0.6 & 0.586/0.58 & 0.62/0.608 \\
\hline

\end{tabular}
\end{minipage}

\begin{minipage}[t][][b]{\textwidth}
\caption{Table showing F1-score/$\kappa$- score for several oversampling strategies (Baseline, SMOTE, Polynom-fit SMOTE, ProWSyn, CURE-SMOTE, SOMO, LoRAS) for $14$ Set-I benchmarking datasets for kNN classifier.}
\label{table_pilot_kNN}
\centering
\tabularnewline
\begin{tabular}{l |@{\hskip3pt}c@{\hskip3pt}|@{\hskip3pt} c @{\hskip3pt}|@{\hskip3pt} c@{\hskip3pt}|@{\hskip3pt} c@{\hskip3pt}| @{\hskip3pt}c @{\hskip3pt}|@{\hskip3pt}c@{\hskip3pt}|@{\hskip3pt}c@{\hskip3pt}}
\hline
     Datasets   &   Baseline &      SMOTE & Polynom-fit SMOTE &    ProWSyn &       CURE &       SOMO &     LORAS \\
\hline
\hline

abalone9-18 & 0.125/0.119 & 0.345/0.286 & 0.368/0.345 & 0.404/0.356 & 0.332/0.279 & 0.125/0.119 & 0.394/0.346 \\
\hline
abalone\_17\_vs\_7\_8\_9\_10 & 0.11/0.105 & 0.295/0.267 & 0.316/0.3 & 0.335/0.309 & 0.286/0.259 & 0.11/0.105 & 0.298/0.271 \\
\hline
car-vgood & 0.594/0.585 & 0.88/0.875 & 0.84/0.833 & 0.836/0.828 & 0.84/0.834 & 0.583/0.574 & 0.898/0.894 \\
\hline
car\_good & 0.409/0.399 & 0.857/0.851 & 0.572/0.546 & 0.509/0.478 & 0.737/0.725 & 0.423/0.413 & 0.84/0.832 \\
\hline
flare-F & 0.095/0.079 & 0.28/0.234 & 0.28/0.234 & 0.275/0.227 & 0.305/0.265 & 0.1/0.083 & 0.293/0.251 \\
\hline
hypothyroid & 0.61/0.595 & 0.589/0.562 & 0.646/0.625 & 0.578/0.551 & 0.668/0.651 & 0.61/0.595 & 0.619/0.595 \\
\hline
kddcup-guess\_passwd\_vs\_satan & 0.99/0.99 & 0.99/0.99 & 0.99/0.99 & 0.99/0.99 & 0.99/0.99 & 0.99/0.99 & 0.99/0.99 \\
\hline
kr-vs-k-three\_vs\_eleven & 0.936/0.935 & 0.949/0.948 & 0.949/0.948 & 0.936/0.934 & 0.937/0.935 & 0.939/0.937 & 0.945/0.943 \\
\hline
kr-vs-k-zero-one\_vs\_draw & 0.892/0.889 & 0.879/0.873 & 0.904/0.9 & 0.879/0.874 & 0.92/0.917 & 0.898/0.895 & 0.91/0.906 \\
\hline
shuttle-2\_vs\_5 & 0.998/0.998 &  1.0/1.0 &  1.0/1.0 & 0.992/0.991 &  1.0/1.0 & 0.998/0.998 &  1.0/1.0 \\
\hline
winequality-red-4 & 0.0/-0.001 & 0.066/0.009 & 0.054/-0.005 & 0.074/0.015 & 0.073/0.016 & 0.0/-0.001 & 0.078/0.023 \\
\hline
yeast4 & 0.139/0.126 & 0.323/0.285 & 0.329/0.292 & 0.292/0.25 & 0.357/0.325 & 0.139/0.126 & 0.383/0.353 \\
\hline
yeast5 & 0.679/0.67 & 0.67/0.656 & 0.67/0.657 & 0.627/0.611 & 0.688/0.676 & 0.679/0.67 & 0.686/0.673 \\
\hline
yeast6 & 0.562/0.553 & 0.332/0.306 & 0.367/0.343 & 0.307/0.28 & 0.49/0.474 & 0.562/0.553 & 0.361/0.338 \\
\hline
Average & 0.51/0.503 & 0.604/0.582 & 0.592/0.572 & 0.574/0.55 & 0.616/0.596 & 0.511/0.504 & 0.621/0.601 \\
\hline

\end{tabular}
\end{minipage}

\begin{minipage}[t][][b]{\textwidth}
\caption{Table showing F1-score/$\kappa$- score for several oversampling strategies (Baseline, SMOTE, Polynom-fit SMOTE, ProWSyn, CURE-SMOTE, SOMO, LoRAS) for $14$ Set-I benchmarking datasets for LR classifier.}
\label{table_pilot_LR}
\centering
\tabularnewline
\begin{tabular}{l |@{\hskip3pt}c@{\hskip3pt}|@{\hskip3pt} c @{\hskip3pt}|@{\hskip3pt} c@{\hskip3pt}|@{\hskip3pt} c@{\hskip3pt}| @{\hskip3pt}c @{\hskip3pt}|@{\hskip3pt}c@{\hskip3pt}|@{\hskip3pt}c@{\hskip3pt}}
\hline
     Datasets   &   Baseline &      SMOTE & Polynom-fit SMOTE &    ProWSyn &       CURE &       SOMO &     LORAS \\
\hline
\hline
abalone9-18 & 0.559/0.694 & 0.524/0.485 & 0.57/0.538 & 0.518/0.477 & 0.563/0.529 & 0.559/0.538 & 0.569/0.536 \\
\hline
abalone\_17\_vs\_7\_8\_9\_10 & 0.357/0.436 & 0.307/0.278 & 0.363/0.339 & 0.323/0.295 & 0.342/0.315 & 0.357/0.347 & 0.324/0.297 \\
\hline
car-vgood & 0.122/0.353 & 0.377/0.337 & 0.397/0.36 & 0.381/0.342 & 0.368/0.327 & 0.132/0.117 & 0.385/0.346 \\
\hline
car\_good & 0.0/0.078 & 0.095/0.024 & 0.103/0.033 & 0.099/0.028 & 0.108/0.039 & 0.022/0.012 & 0.094/0.024 \\
\hline
flare-F & 0.208/0.352 & 0.263/0.212 & 0.329/0.287 & 0.265/0.214 & 0.273/0.224 & 0.252/0.23 & 0.28/0.23 \\
\hline
hypothyroid & 0.335/0.444 & 0.368/0.317 & 0.401/0.355 & 0.381/0.331 & 0.388/0.343 & 0.335/0.316 & 0.376/0.326 \\
\hline
kddcup-guess\_passwd\_vs\_satan & 0.996/0.989 &  1.0/1.0 & 0.996/0.996 &  1.0/1.0 &  1.0/1.0 &  1.0/1.0 &  1.0/1.0 \\
\hline
kr-vs-k-three\_vs\_eleven & 0.96/0.997 & 0.947/0.946 & 0.948/0.947 & 0.953/0.951 & 0.949/0.948 & 0.96/0.959 & 0.946/0.944 \\
\hline
kr-vs-k-zero-one\_vs\_draw & 0.853/0.934 & 0.732/0.719 & 0.795/0.786 & 0.74/0.728 & 0.766/0.756 & 0.862/0.857 & 0.764/0.753 \\
\hline
shuttle-2\_vs\_5 &  1.0/1.0 &  1.0/1.0 & 0.983/0.983 &  1.0/1.0 &  1.0/1.0 &  1.0/1.0 &  1.0/1.0 \\
\hline
winequality-red-4 & 0.007/0.155 & 0.127/0.072 & 0.132/0.078 & 0.129/0.074 & 0.157/0.106 & 0.007/0.004 & 0.138/0.084 \\
\hline
yeast4 & 0.212/0.415 & 0.255/0.211 & 0.275/0.233 & 0.259/0.215 & 0.247/0.203 & 0.212/0.2 & 0.26/0.217 \\
\hline
yeast5 & 0.552/0.709 & 0.599/0.582 & 0.631/0.616 & 0.607/0.59 & 0.622/0.607 & 0.552/0.541 & 0.598/0.581 \\
\hline
yeast6 & 0.437/0.512 & 0.3/0.272 & 0.345/0.32 & 0.295/0.266 & 0.398/0.377 & 0.437/0.428 & 0.323/0.296 \\
\hline
Average & 0.471/0.576 & 0.492/0.461 & 0.519/0.491 & 0.496/0.465 & 0.513/0.484 & 0.478/0.468 & 0.504/0.474 \\
\hline

\end{tabular}
\end{minipage}
\end{table*}

\end{document}